\documentclass[runningheads]{llncs}

\usepackage{eccv}

\usepackage{eccvabbrv}

\usepackage{graphicx}
\usepackage{booktabs}
\usepackage{enumitem}
\usepackage{gensymb}
\usepackage{float}
\usepackage{makecell}
\usepackage{multirow}
\usepackage{wrapfig}

\usepackage[accsupp]{axessibility}  

\usepackage[pagebackref,breaklinks,colorlinks,citecolor=eccvblue]{hyperref}

\usepackage{orcidlink}

\definecolor{cvprblue}{rgb}{0.21,0.49,0.74}

\DeclareMathAlphabet{\mymathbb}{U}{BOONDOX-ds}{m}{n}

\definecolor{azure}{rgb}{0.0, 0.5, 1.0}
\definecolor{myanchor}{RGB}{0,0,0}
\definecolor{mypositive}{RGB}{23,255,49}
\definecolor{mynegative}{RGB}{209,26,66}

\DeclareCaptionLabelFormat{andtable}{#1~#2  \&  \tablename~\thetable}

\begin{document}

\title{Dissolving Is Amplifying: Towards Fine-Grained Anomaly Detection
} 

\titlerunning{Dissolving Is Amplifying}

\author{
    Jian Shi\inst{1}
    \and  Pengyi Zhang\inst{2}
    \and  Ni Zhang\inst{2}
    \and  Hakim Ghazzai\inst{1}
    \and  Peter Wonka\inst{1}
}

\authorrunning{J. Shi et al.}

\institute{King Abdullah University of Science and Technology, Thuwal, Saudi Arabia
\email{\{jian.shi, hakim.ghazzai, peter.wonka\}@kaust.edu.sa}\\ \and
NEC Laboratories China, Beijing, China\\
\email{\{zhang\_pengyi,zhangni\_nlc\}@nec.cn}
}

\maketitle

\begin{abstract}
Medical imaging often contains critical fine-grained features, such as tumors or hemorrhages, which are crucial for diagnosis yet potentially too subtle for detection with conventional methods. 
In this paper, we introduce \textit{DIA}, dissolving is amplifying. DIA is a fine-grained anomaly detection framework for medical images.
First, we introduce \textit{dissolving transformations}.
We employ diffusion with a generative diffusion model as a dedicated feature-aware denoiser. Applying diffusion to medical images in a certain manner can remove or diminish fine-grained discriminative features.
Second, we introduce an \textit{amplifying framework} based on contrastive learning to learn a semantically meaningful representation of medical images in a self-supervised manner, with a focus on fine-grained features. The amplifying framework contrasts additional pairs of images with and without dissolving transformations applied and thereby emphasizes the dissolved fine-grained features.
DIA significantly improves the medical anomaly detection performance with around 18.40\% AUC boost against the baseline method and achieves an overall SOTA against other benchmark methods. Our code is available at
\url{https://github.com/shijianjian/DIA.git}.
\end{abstract}

\section{Introduction}
\label{sec:intro}
Anomaly detection aims to detect exceptional data instances that significantly deviate from normal data. A popular application is the detection of anomalies in medical images, where these anomalies often indicate a form of disease or medical problem. In the medical field, anomalous data is scarce and diverse, so anomaly detection is commonly modeled as semi-supervised anomaly detection. This means that anomalous data is not available during training, and the training data contains only the "normal'' class.\footnote[1]{\scriptsize Some early studies refer to training with only normal data as unsupervised anomaly detection. However, we follow~\cite{Musa2021,Pang2021-ch} and other newer methods and use the term semi-supervised.}
Traditional anomaly detection methods include one-class methods (\eg One-class SVM~\cite{chen2001one}), reconstruction-based methods (\eg AutoEncoders~\cite{williams2002comparative}), and statistical models (\eg HBOS~\cite{goldstein2012histogram}). However, most anomaly detection methods suffer from a low recall rate, meaning that many normal samples are wrongly reported as anomalies while true yet sophisticated anomalies are missed~\cite{Pang2021-ch}.
Notably, due to the nature of anomalies, the collection of anomaly data can hardly cover all anomaly types, even for supervised classification-based methods \cite{deepweak1910.13601}.
An inherited challenge is the inconsistent behavior of anomalies, which varies without a concrete definition~\cite{thudumu2020comprehensive,chalapathy2019deep}.
Thus, identifying unseen anomalous features without requiring prior knowledge of anomalous feature patterns is crucial to anomaly detection applications.




In order to identify unseen anomalous features, many studies leveraged data augmentations~\cite{GeoTrans1805.10917,Ye2022} and adversarial features~\cite{Akcay2019-tm} to emphasize various feature patterns that deviate from normal data. This field attracted more attention after incorporating Generative Adversarial Networks (GANs)~\cite{NIPS2014_GAN}, including~\cite{sabokrou2018adversarially,pmlr-v80-ruff18a,anogan-1703.05921,Akcay2019-jx,Akcay2019-tm,Zhao2018,DeepDisa-2202.00050}, to enlarge the feature distances between normal and anomalous features through adversarial data generation methods.
Furthermore, some studies~\cite{salem2018anomaly,pourreza2021g2d,murase2022algan} explored the use of GANs to deconstruct images to generate out-of-distribution data for obtaining more varied anomalous features.
Inspired by the recent successes of contrastive learning~\cite{chen2020simple,chen2020big,he2019moco,chen2020mocov2,grill2020bootstrap,chen2020exploring,caron2020unsupervised}, contrastive-based anomaly detection methods such as Contrasting Shifted Instances (CSI)~\cite{tack2020csi} and mean-shifted contrastive loss~\cite{reiss2021mean} improve upon GAN-based methods by a large margin.
The contrastive-based methods fit the anomaly detection context well, as they are able to learn robust feature encoding without supervision. By comparing the feature differences between positive pairs (\eg the same image with different views) and negative pairs (\eg different images w/wo different views) without knowing the anomalous patterns, contrastive-based methods achieved outstanding performance in many general anomaly detection tasks \cite{tack2020csi,reiss2021mean}.
However, given the low performance in experiments in~\cref{sec:exp}, those methods are less effective for medical anomaly detection.
We suspect that contrastive learning in conjunction with traditional data augmentation methods (\eg crop, rotation) cannot focus on fine-grained features and only identifies coarse-grained feature differences well (\eg car vs. plane).
As a result, medical anomaly detection remains challenging because models struggle to recognize these fine-grained, inconspicuous, yet important anomalous features that manifest differently across individual cases. These features are critical for identifying anomalies but can be subtle and easily overlooked.
Thus, in this work, we investigate the principled question: \textit{how to emphasize the fine-grained features for fine-grained anomaly detection?}

\begin{figure*}[!t]
\centering
\begin{subfigure}[t]{.192\linewidth}
  \centering
  \includegraphics[width=.98\linewidth,trim={0 9.1cm 0 0}, clip]{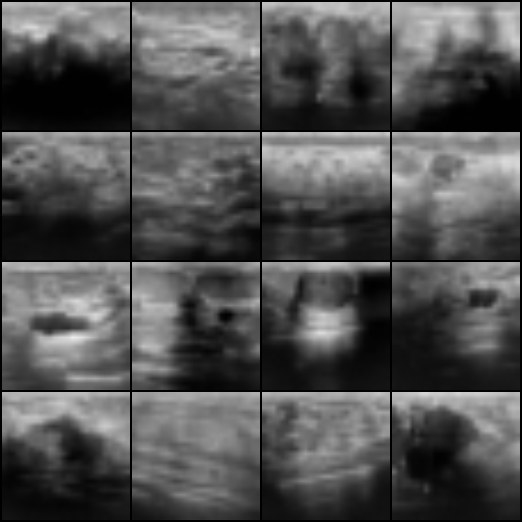}
  \includegraphics[width=.98\linewidth,trim={0 9.1cm 0 0}, clip]{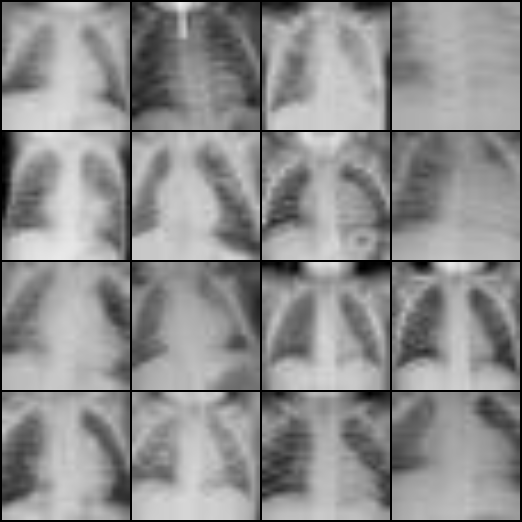}
  \includegraphics[width=.98\linewidth,trim={0 9.1cm 0 0}, clip]{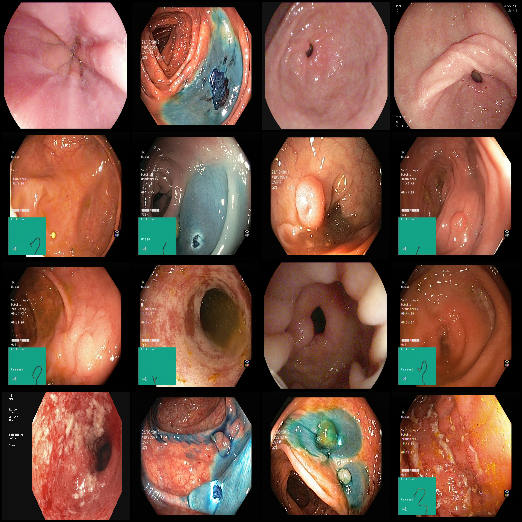}
  \includegraphics[width=.98\linewidth,trim={0 9.1cm 0 0}, clip]{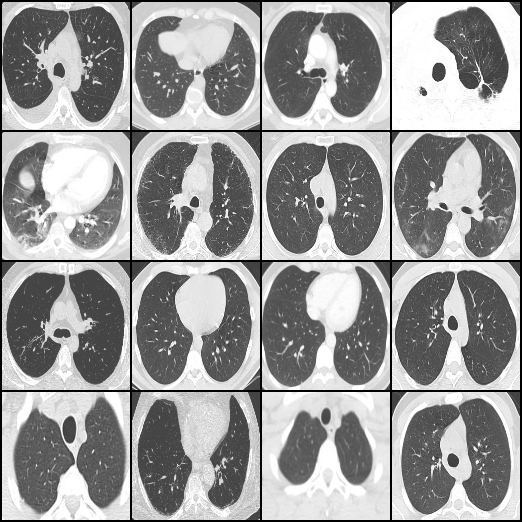}
  \caption{Input Images}
  \label{fig:sfig1}
\end{subfigure}%
\begin{subfigure}[t]{.192\linewidth}
  \centering
  \includegraphics[width=.98\linewidth,trim={0 9.1cm 0 0}, clip]{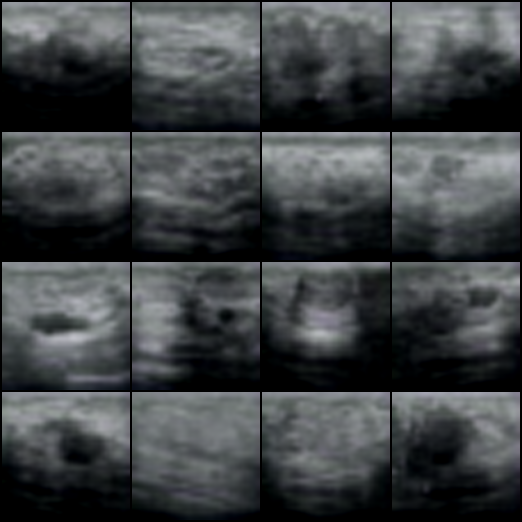}
  \includegraphics[width=.98\linewidth,trim={0 9.1cm 0 0}, clip]{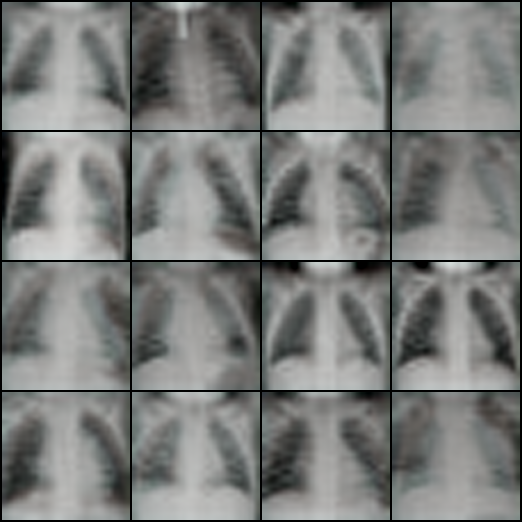}
  \includegraphics[width=.98\linewidth,trim={0 9.1cm 0 0}, clip]{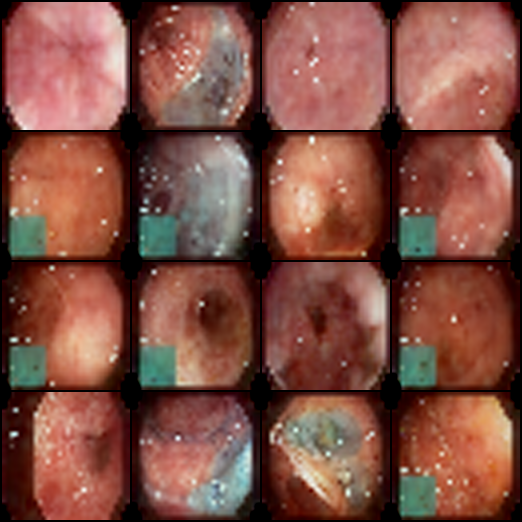}
  \includegraphics[width=.98\linewidth,trim={0 9.1cm 0 0}, clip]{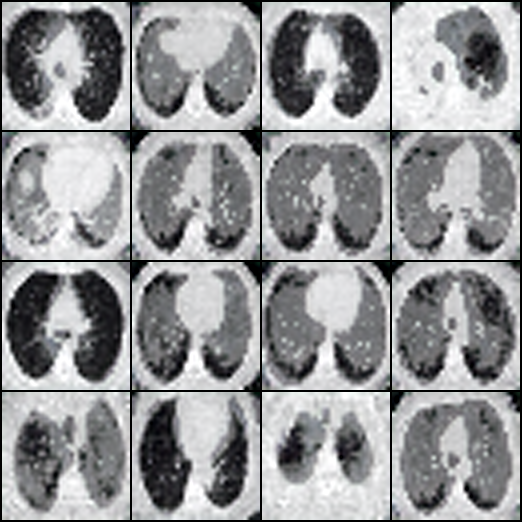}
  \caption{$t=50$}
  \label{fig:sfig1.5}
\end{subfigure}%
\begin{subfigure}[t]{.192\linewidth}
  \centering
  \includegraphics[width=.98\linewidth,trim={0 9.1cm 0 0}, clip]{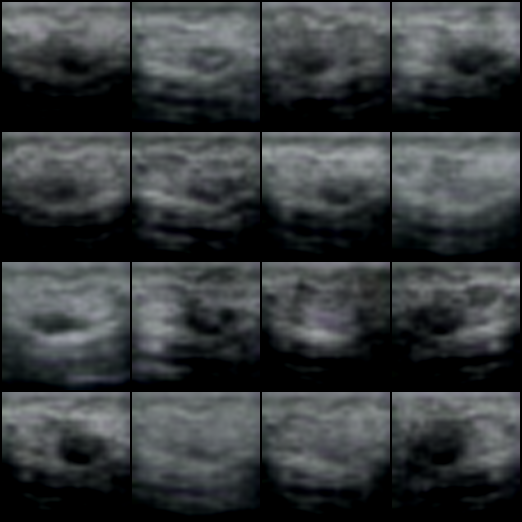}
  \includegraphics[width=.98\linewidth,trim={0 9.1cm 0 0}, clip]{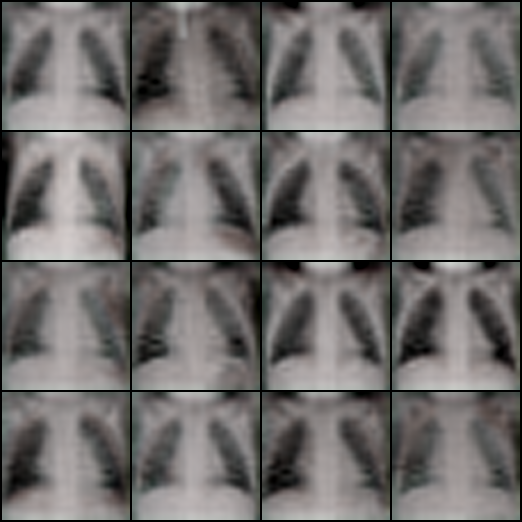}
  \includegraphics[width=.98\linewidth,trim={0 9.1cm 0 0}, clip]{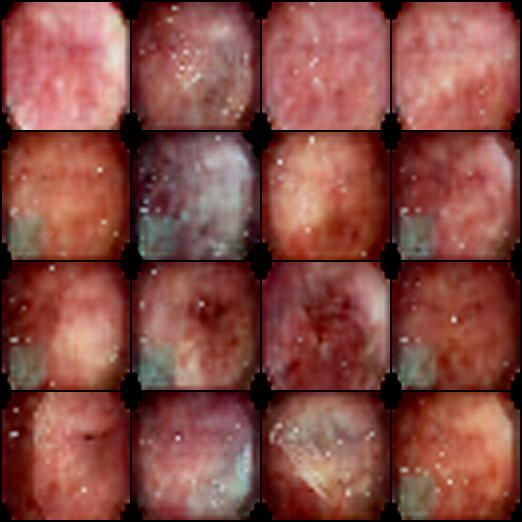}
  \includegraphics[width=.98\linewidth,trim={0 9.1cm 0 0}, clip]{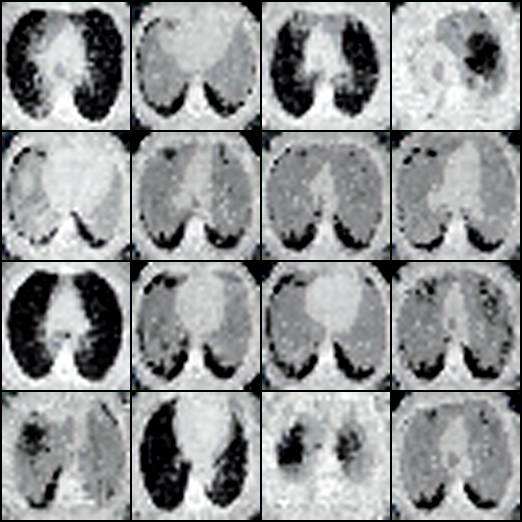}
  \caption{$t=100$}
  \label{fig:sfig2}
\end{subfigure}%
\begin{subfigure}[t]{.192\linewidth}
  \centering
  \includegraphics[width=.98\linewidth,trim={0 9.1cm 0 0}, clip]{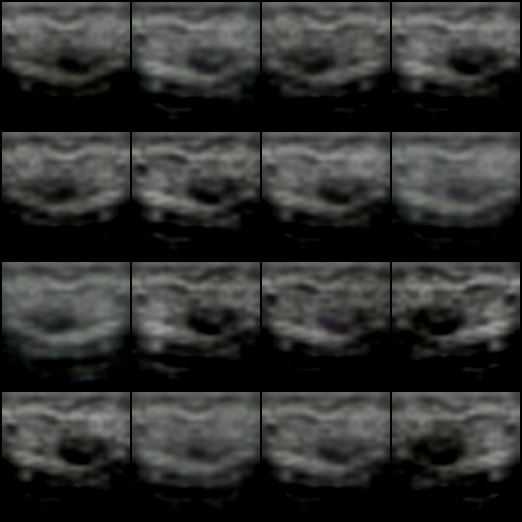}
  \includegraphics[width=.98\linewidth,trim={0 9.1cm 0 0}, clip]{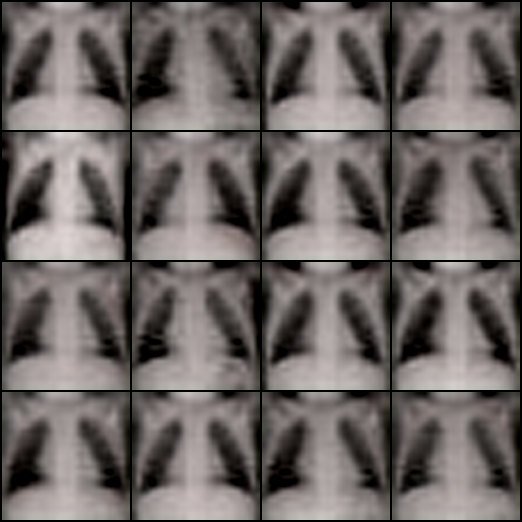}
  \includegraphics[width=.98\linewidth,trim={0 9.1cm 0 0}, clip]{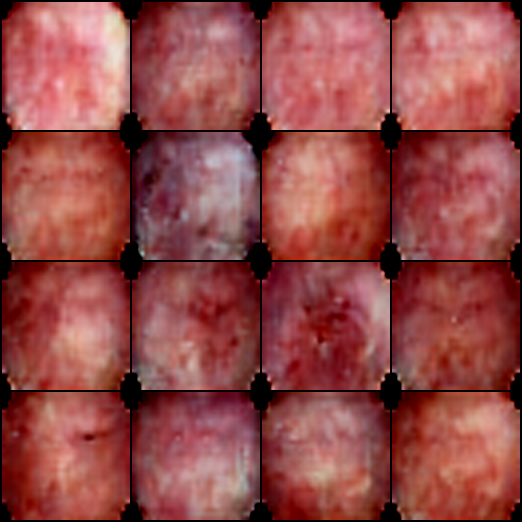}
  \includegraphics[width=.98\linewidth,trim={0 9.1cm 0 0}, clip]{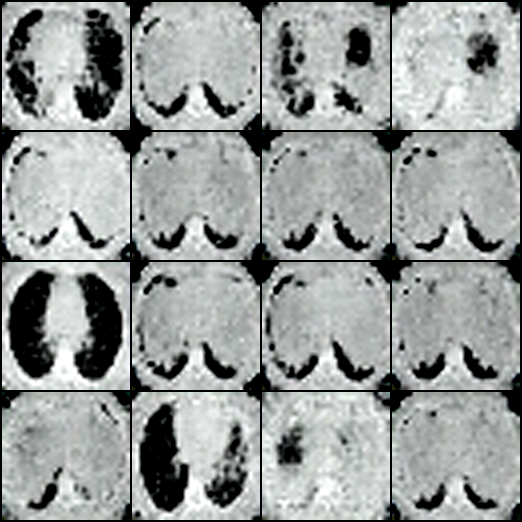}
  \caption{$t=200$}
  \label{fig:sfig3}
\end{subfigure}%
\begin{subfigure}[t]{.192\linewidth}
  \centering
  \includegraphics[width=.98\linewidth,trim={0 9.1cm 0 0}, clip]{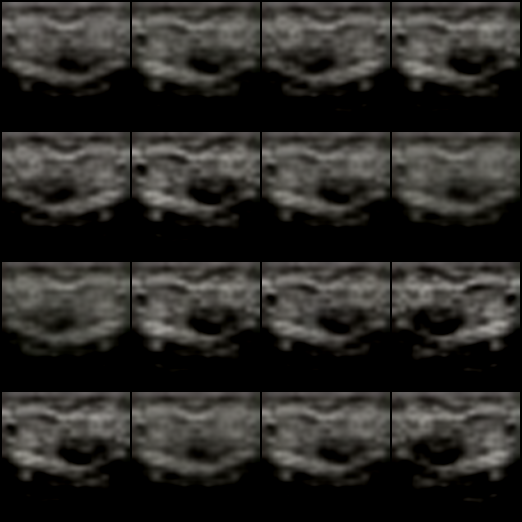}
  \includegraphics[width=.98\linewidth,trim={0 9.1cm 0 0}, clip]{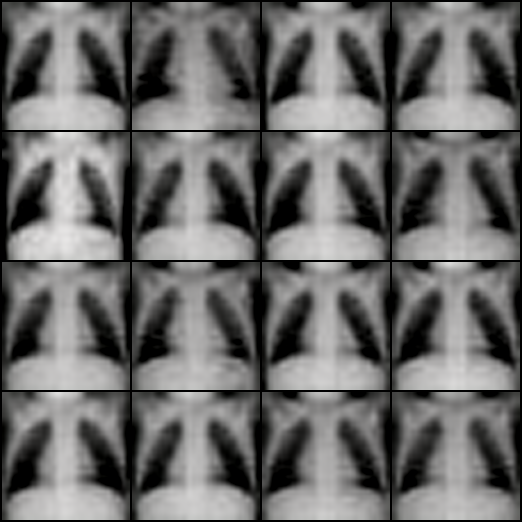}
  \includegraphics[width=.98\linewidth,trim={0 9.1cm 0 0}, clip]{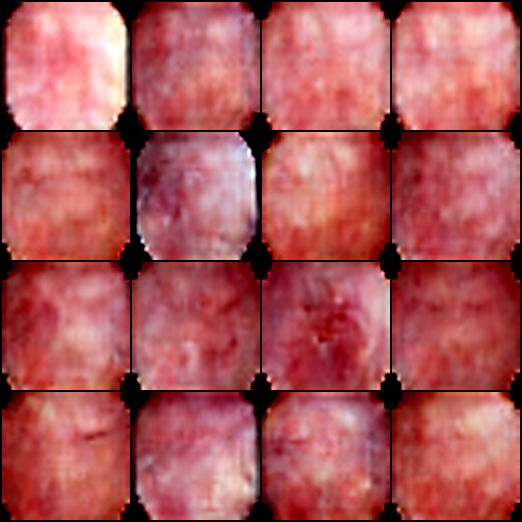}
  \includegraphics[width=.98\linewidth,trim={0 9.1cm 0 0}, clip]{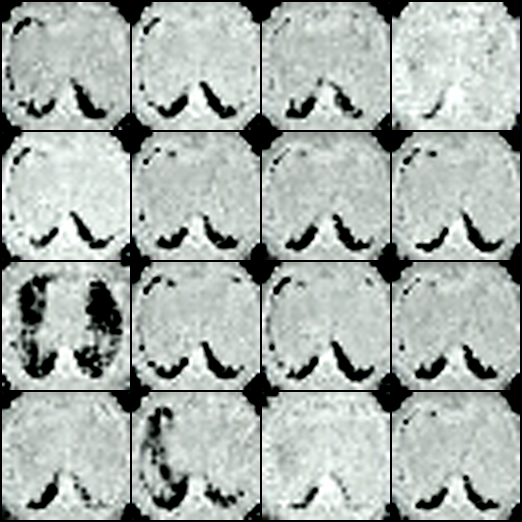}
  \caption{$t=400$}
  \label{fig:sfig4}
\end{subfigure}%
\setlength{\belowcaptionskip}{-2em}
\setlength{\abovecaptionskip}{0em}
\caption{Dissolving Transformations. \cref{fig:sfig1.5,fig:sfig2,fig:sfig3,fig:sfig4} show how the fine-grained features are dissolved (removed or suppressed). This effect is stronger as the time step $t$ is increased from left to right. In the extreme case, in \cref{fig:sfig4}, different input images become very similar or almost identical depending on the dataset. We show results for four datasets from top to bottom.}
\label{fig:demo}
\end{figure*}

\textbf{Our method.}
This paper dissects the complex feature patterns within medical datasets into two distinct categories: discriminative and non-discriminative features. Discriminative features are commonly unique and fine-grained characteristics that allow for the differentiation of individual data samples, serving as critical markers for identification and classification. Conversely, non-discriminative features encompass the shared patterns that define the general semantic context of the dataset, offering a backdrop against which the discriminative features stand out. To aid the learning of fine-grained discriminative feature patterns, we propose an intuitive contrastive learning strategy to compare an image against its transformed version with fewer discriminative features to emphasize the removed fine-grained details.
We introduce \textit{dissolving transformations} based on pre-trained diffusion models, that leverage the individual reverse diffusion steps within the diffusion models to function as feature-aware denoisers, to remove or suppress fine-grained discriminative features from an input image.
We also introduce the framework \textit{DIA}, dissolving is amplifying, that leverages the proposed \textit{dissolving transformations}. DIA is a contrasting learning framework. Its enhanced understanding of fine-grained discriminative features stems from a loss function that contrasts images that have been transformed with dissolving transformations to images that have not. On six medical datasets, our method obtained roughly an 18.40\% AUC boost against the baseline method and achieved the overall SOTA compared to existing methods for fine-grained medical anomaly detection.
Key contributions of \textit{DIA} include:
\begin{itemize}[leftmargin=1em]
    \setlength\itemsep{0.em}
    \item \textbf{Conceptual Contribution.} We propose a novel strategy that enhances the detection of fine-grained, subtle anomalies without requiring pre-defined anomalous feature patterns, by emphasizing the differences between images and their feature-dissolved counterparts.
    \item \textbf{Technical Contribution 1.} We introduce \textit{dissolving transformations} to dissolve the fine-grained features of images. It performs semantic feature dissolving via the reverse process of diffusion models as described in \cref{fig:demo}.
    \item \textbf{Technical Contribution 2.} We present an \textit{amplifying strategy} for self-supervised fine-grained feature learning, leveraging a \textit{fine-grained NT-Xent loss} to learn fine-grained discriminative features.
\end{itemize}

\section{Related Work}

\subsection{Synthesis-based Anomaly Detection} As ~\cite{Pang2021-ch,BeulaRani2020} indicated, semi-supervised anomaly detection methods dominated this research field. These methods utilized only normal data whilst training. With the introduction of GANs~\cite{NIPS2014_GAN}, many attempts have been made to bring GANs into anomaly detection. Here, we roughly categorize current methods to \textit{reconstructive synthesis} that increases the variation of normal data, and \textit{deconstructive synthesis} that generates more anomalous data.

\noindent\textbf{Reconstructive Synthesis.} Many studies~\cite{brock2018large,zhang21datasetgan} focused on synthesizing various in-distribution data (\ie normal data) with synthetic methods. For anomaly detection tasks, earlier works such as AnoGAN~\cite{Schlegl2017-zr} learn normal data distributions with GANs that attempt to reconstruct the most similar images by optimizing a latent noise vector iteratively. With the success of Adversarial Auto Encoders (AAE)~\cite{AAE2016}, some more recent studies combined AutoEncoders and GANs together to detect anomalies. GANomaly~\cite{Akcay2019-jx} further regularized the latent spaces between inputs and reconstructed images, and then some following works improved it with more advanced generators such as UNet~\cite{Akcay2019-tm} and UNet++~\cite{Cheng2020}. AnoDDPM~\cite{Wyatt_2022_CVPR} replaced GANs with diffusion model generators and stated the effectiveness of noise types for medical images (i.e., Simplex noise is better than Gaussian noise). In general, most of the \textit{reconstructive synthesis} methods aim to improve normality feature learning despite the awareness of abnormalities, which impedes the model from understanding the anomaly feature patterns.

\noindent\textbf{Deconstructive Synthesis.} Due to the difficulties of data acquisition and to protect patient privacy, getting high-quality, balanced datasets in the medical field is difficult~\cite{ker2017deep}. Thus, \textit{deconstructive synthesis} methods are widely applied in medical image domains, such as X-ray~\cite{salehinejad2018generalization}, lesion~\cite{frid2018gan}, and MRI~\cite{han2018gan}. Recent studies tried to integrate such negative data generation methods into anomaly detection. G2D~\cite{pourreza2021g2d} proposed a two-phased training to train an anomaly image generator and then an anomaly detector. Similarly, ALGAN~\cite{murase2022algan} proposed an end-to-end method that generates \textit{pseudo-anomalies} during the training of anomaly detectors. Such GAN-based methods deconstruct images to generate \textit{pseudo-anomalies}, resulting in unrealistic anomaly patterns, though multiple regularizers are applied to preserve image semantics.
Unlike most works to synthesize novel samples from noises, we dissolve the fine-grained features on input data. Our method, therefore, learns the fine-grained instance feature patterns by comparing samples against their feature-dissolved counterparts. Benefiting from the step-by-step diffusing process of diffusion models, our proposed \textit{dissolving transformations} can provide fine control over feature dissolving levels.

\subsection{Contrastive-based Anomaly Detection} To improve anomaly detection performances, previous studies such as~\cite{Dosovitskiy2014Discriminative,wen2016discriminative} explored the discriminative feature learning to reduce the needs of labeled samples for supervised anomaly detection. More recently, GeoTrans~\cite{GeoTrans1805.10917} leveraged geometric transformations to learn discriminative features, which significantly improved anomaly detection abilities. ARNet~\cite{Ye2022} attempted to use embedding-guided feature restoration to learn more semantic-preserving anomaly features.
Specifically, contrastive learning methods~\cite{chen2020simple,chen2020big,he2019moco,chen2020mocov2,grill2020bootstrap,chen2020exploring,caron2020unsupervised} are proven to be promising in unsupervised representation learning. Inspired by the recent integration~\cite{tack2020csi,reiss2021mean,Cho2021} of contrastive learning and anomaly detection tasks, we propose to construct negative pairs of a given sample and its corresponding \textit{feature-dissolved} samples in a contrastive manner to enhance the awareness of fine-grained discriminative features for medical anomaly detection.

\begin{figure*}[t!]
    \centering
    \includegraphics[width=1\linewidth]{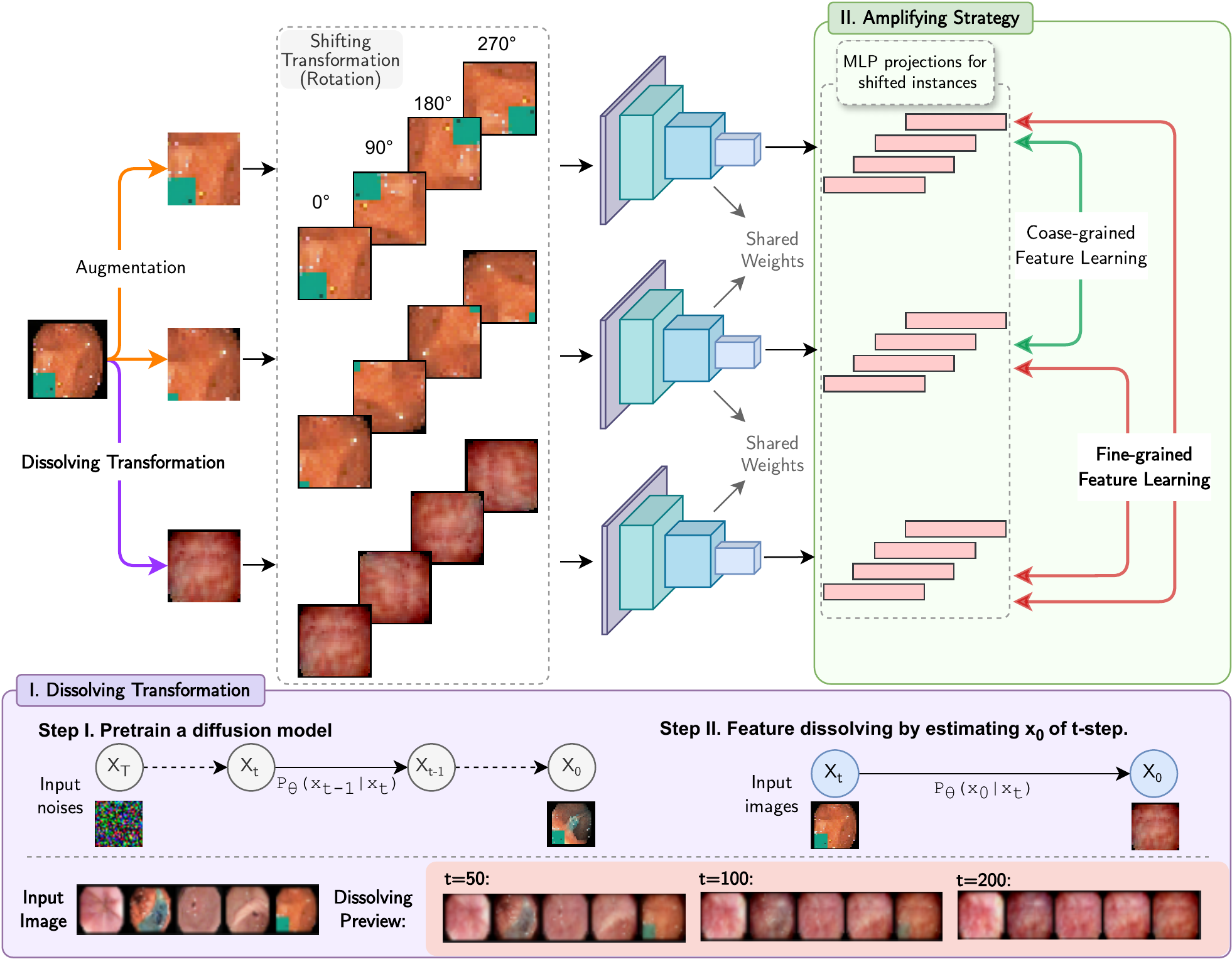}
    \setlength{\belowcaptionskip}{-1.5em}
    \setlength{\abovecaptionskip}{-.5em}
    \caption{An overview of the DIA framework as applied to the Kvasir-polyp dataset. (I) With a pretrained diffusion model, we perform feature-aware dissolving transformations on an image $x$. This process estimates the denoised version $x_0$ of $x$ at a given time step $t$, resulting in a feature-dissolved image $\hat{x}$. As $t$ increases, $\hat{x}$ progressively loses its fine-grained discriminative features, highlighting the dissolving effect of removing discriminative image features. (II) Given images, we generate transformed versions with augmentations and dissolving transformation. We form positive and negative pairs as described in~\cref{sec:con_loss}. Our framework particularly learns fine-grained features by contrasting between original images and their feature-dissolved counterparts.
    }
    \label{fig:diffusion}
\end{figure*}

\section{Methodology}

This section introduces DIA (Dissolving Is Amplifying), a method curated for fine-grained anomaly detection for medical imaging.
DIA is a self-supervised method based on contrastive learning, as illustrated in~\cref{fig:diffusion}. DIA learns representations that can distinguish fine-grained discriminative features in medical images.
First, DIA employs a dissolving strategy based on \textit{dissolving transformations} (\cref{sec:dissolver}). The dissolving transformations can remove or deemphasize fine-grained discriminative features. Second, DIA uses the amplifying framework described in \cref{sec:amp} to contrast images that have been transformed with and without dissolving transformations. We use the term amplifying framework as it amplifies the representation of fine-grained discriminative features.

\subsection{Dissolving Strategy}
\label{sec:dissolver}

We introduce \textit{dissolving transformations} to create negative examples in a contrastive learning framework. The dissolving transformations are achieved by pre-trained diffusion models. The output image maintains a similar structure and appearance to the input image, but several fine-grained discriminative features unique to the input image are removed or attenuated.
Unlike the regular diffusion process that starts with pure noise, we initialize with the input image without adding noise.
As depicted in \cref{fig:demo}, \textit{dissolving transformations} progressively remove fine-grained details from various datasets (\cref{fig:sfig1.5,fig:sfig2,fig:sfig3,fig:sfig4}) with increasing diffusion time steps $t$.

To recap, diffusion models consist of forward and reverse processes, each performed over $T$ time steps. The forward process $q$ gradually adds noise to an image $x_0$ for $T$ steps to obtain a pure noise image $x_T$, whereas the reverse process $p$ aims at restoring the starting image $x_0$ from $x_T$. In particular, we sample an image $x_0\sim q(x_0)$ from a real data distribution $q(x_0)$, then add noise at each step $t$ with the forward process $q(x_t|x_{t-1})$, which can be expressed as:
\begin{align}
    \small
    q(x_t|x_{t-1})&=\mathcal{N}(x_t;\sqrt{1-\beta_t}\cdot x_{t-1},\beta_t \cdot \text{I}), \\
    q(x_{1:T}|x_0)&=\prod_{t=1}^T q(x_t|x_{t-1}),
\end{align}
where $\beta_t$ represents a known variance schedule that follows $0<\beta_1<\beta_2<\cdots<\beta_T < 1$. Afterwards, the reverse process removes noise starting at $p(x_T)=\mathcal{N}(x_T;0,\text{I})$ for $T$ steps. Let $\theta$ be the network parameters:
\begin{align}
    \small
    p_{\theta}(x_{t-1} | x_t) = \mathcal{N}(x_{t-1};\mu_{\theta}(x_t, t), \Sigma_\theta(x_t, t)),
    \label{eq:p_sample}
\end{align}
where $\mu_\theta$ and $\Sigma_\theta$ are the mean and variance conditioned on step number $t$.




The proposed \textit{dissolving transformations} are based on \cref{eq:p_sample}.
Instead of generating images by progressive denoising, we apply reverse diffusion in a single step directly on an input image.
Essentially, we set $x_t = x$ in \cref{eq:p_sample}, where $x$ is the input image.
We then compute an approximated state $x_0$ and denote it as $\hat{x}_{t\rightarrow 0}$ to make it clear that the equation below is parameterized by the time step $t$.
By reparametrizing \cref{eq:p_sample}, $\hat{x}_{t\rightarrow 0}$ can be obtained by:
\begin{align}
    \small
    \hat{x}_{t\rightarrow 0}=\sqrt{\frac{1}{\Bar{\alpha}_t}} \cdot x - \sqrt{\frac{1}{\Bar{\alpha}_t} - 1}\cdot \epsilon_\theta(x,t), \quad
    \Bar{\alpha}_t := \Pi_{s=1}^{t} \alpha_s\;\text{and}\;\alpha_t:=1-\beta_t,
\end{align}
where $\epsilon_\theta$ is a function approximator (\eg UNet) to predict the corresponding noise from $x$.
Since a greater value of $t$ leads to a higher variance $\beta_t$, $\hat{x}_{t\rightarrow 0}$ is expected to remove more of the "noise" if $t$ is large.
In our context, we do not remove "noise" but discriminative features. If $t$ is small, the removed discriminative features are more fine-grained. If $t$ is larger, larger discriminative features may be removed. See~\cref{fig:demo} and~\cref{sec:discussion} for examples and an in-depth discussion.

\subsection{Amplifying Framework}
\label{sec:amp}

We propose a novel contrastive learning framework to enhance the awareness of the fine-grained image features by integrating the proposed \textit{dissolving transformations}.
In anomaly detection, the efficacy of contrastive learned features can be enhanced by applying \textit{shifting transformations}~\cite{tack2020csi}. A typical example is using significant rotations, which alters the distribution of the data based on the orientation of the transformed images.
For instance, images rotated by 90 degrees are assimilated into the same distribution, whereas images subjected to a 180-degree rotation diverge from this distribution.
However, this improved contrastive feature learning technique does not come with a fine-grained feature learning mechanism, resulting in low performances on fine-grained anomaly detection tasks.
We introduce feature-dissolved samples to augment the process of fine-grained feature learning.
The feature-dissolved samples present significant differences from the original data, despite both sets belonging to the same shifting distributions.
In particular, we aim to enforce the model to focus on fine-grained features by emphasizing the differences between images with and without \textit{dissolving transformations}. 

In our amplifying framework, we employ three types of transformations: \textit{shifting transformations} (\eg large rotations), \textit{non-shifting transformations} (\eg color jitter, random resized crop, and grayscale), and \textit{dissolving transformations}.
Our contrastive learning framework uniquely applies these transformations to input images through $3K$ distinct processes. The first $2K$ transformation branches are dedicated to coarse-grained feature learning, focusing on broader, more general features of the data. Conversely, the final $K$ transformations are specifically tailored for fine-grained feature learning. This is accomplished by contrasting the transformed images against non-dissolved data samples, thereby enhancing the model's ability to discern subtle differences within the data. This approach not only broadens the scope of feature extraction but also significantly improves the model's precision in identifying nuanced patterns and anomalies.



\subsubsection{Transformation Branches}
\label{sec:form}


We use a set $\mathcal{S}$ of $K$ different \textit{shifting transformations}. This set contains only fixed (non-random) transformations and starts from the identity $I$ so that $\mathcal{S} :=\{S_0=I,S_1,\dots,S_{K-1}\}$. 
With input image $x$, we obtain $S_1(x),\dots, S_{K-1}(x)$ as shifted images that strongly differ from the in-distribution samples $S_0(x)=x$. Each of these $K$ shifted images then passes through multiple non-shifting transformations $\in \mathcal{T}$. This yields the set of combined transformations $O:=\{O_0,O_1,\dots,O_{K-1}\} \;\text{and}\; O_k=\mathcal{T} \circ S_k$.
With a slight abuse of notations, we use $\mathcal{T}$ as a sequence of random non-shifting transformations.
This process is then repeated a second time, yielding another transformation set $\mathcal{O}'$. 
We also refer to $\mathcal{O}$ and $\mathcal{O}'$ as two augmentation branches.
Each image is therefore transformed $2K$ times, $K$ times in each augmentation branch. All transformations have supposedly different randomly sampled non-shifting transformations, but $O_i(x)$ and $O'_j(x)$ share the same \textit{shifting transformation} if $i=j$.
The introduced \textit{dissolving transformations} serves as the third augmentation branch, denoted as $\mathcal{A}:=\{A_{0},\dots, A_{K-1}\}$. The \textit{dissolving transformations} branch outputs transformations of the form:
\begin{equation}
    \small
    {A}_k=  \mathcal{T} \circ S_k \circ \mathcal{D}
\end{equation}
where $\mathcal{T}$ is a sequence of random non-shifting transformations, $S_k$ is a \textit{shifting transformation}, and $\mathcal{D}$ is a randomly sampled \textit{dissolving transformation}. In summary, this yields $3K$ transformations of each image, $K$ in each of the three augmentation branches.



\subsubsection{Fine-grained Contrastive Learning}
\label{sec:con_loss}

The goal of contrastive learning is to transform input images into a semantically meaningful feature representation.
It is achieved by bringing similar examples (\ie \textit{positive pairs}) closer and pushing dissimilar examples (\ie \textit{negative pairs}) apart.
To emphasize fine-grained features, an inherent strategy is to create negative pairs, where an image is contrasted with its transformed version with less fine-grained details, thereby enhancing the model's focus on these subtle distinctions.

\begin{wrapfigure}{r}{0.49\textwidth}
\centering
\begin{subfigure}[b]{\linewidth}
\centering
\vspace{-2.em}
\begin{tikzpicture}[scale=0.38]
  \foreach \y [count=\n] in {
      {100,70,70,70,0,70, 70,70,70,70,70,70},
      {70,100,70,70,70,0, 70,70,70,70,70,70},
      {70,70,100,70,70,70, 0,70,70,70,70,70},
      {70,70,70,100,70,70, 70,0,70,70,70,70},
      {0,70,70,70,100,70, 70,70,70,70,70,70},
      {70,0,70,70,70,100, 70,70,70,70,70,70},
      {70,70,0,70,70,70, 100,70,70,70,70,70},
      {70,70,70,0,70,70, 70,100,70,70,70,70},
      {70,70,70,70,70,70, 70,70,100,70,70,70},
      {70,70,70,70,70,70, 70,70,70,100,70,70},
      {70,70,70,70,70,70, 70,70,70,70,100,70},
      {70,70,70,70,70,70, 70,70,70,70,70,100},
    } {
      \ifnum \n=1
        \node[minimum size=4mm,rotate=-90] at (\n, 1) {\scriptsize $O_0(x_1)$};
      \fi
      \ifnum \n=2
        \node[minimum size=4mm,rotate=-90] at (\n, 1) {\scriptsize $O_0(x_2)$};
      \fi
      \ifnum \n=3
        \node[minimum size=4mm,rotate=-90] at (\n, 1) {\scriptsize $O_1(x_1)$};
      \fi
      \ifnum \n=4
        \node[minimum size=4mm,rotate=-90] at (\n, 1) {\scriptsize $O_1(x_2)$};
      \fi
      \ifnum \n=5
        \node[minimum size=4mm,rotate=-90] at (\n, 1) {\scriptsize $O'_0(x_1)$};
      \fi
      \ifnum \n=6
        \node[minimum size=4mm,rotate=-90] at (\n, 1) {\scriptsize $O'_0(x_2)$};
      \fi
      \ifnum \n=7
        \node[minimum size=4mm,rotate=-90] at (\n, 1) {\scriptsize $O'_1(x_1)$};
      \fi
      \ifnum \n=8
        \node[minimum size=4mm,rotate=-90] at (\n, 1) {\scriptsize $O'_1(x_2)$};
      \fi
      \ifnum \n=9
        \node[minimum size=4mm,rotate=-90] at (\n, 1) {\scriptsize ${A_0(x_1,t)}$};
      \fi
      \ifnum \n=10
        \node[minimum size=4mm,rotate=-90] at (\n, 1) {\scriptsize ${A_0(x_2,t)}$};
      \fi
      \ifnum \n=11
        \node[minimum size=4mm,rotate=-90] at (\n, 1) {\scriptsize ${A_1(x_1,t)}$};
      \fi
      \ifnum \n=12
        \node[minimum size=4mm,rotate=-90] at (\n, 1) {\scriptsize ${A_1(x_2,t)}$};
      \fi
      \foreach \x [count=\m] in \y {
        \node[fill=white!\x!blue, draw=black!20, ultra thin, minimum size=4mm, text=white] at (\m,-\n) {};
      }
    }

  \draw[red, very thick, fill=red!10, opacity=0.3]
    (1 - 0.5, -9 + 0.5) --
    (1 - 0.5, -12 - 0.5) --
    (12 + 0.5, -12 - 0.5) --
    (12 + 0.5, -1 + 0.5) --
    (9 - 0.5, -1 + 0.5) --
    (9 - 0.5, -9 + 0.5) -- cycle;
  \foreach \a [count=\i] in {
    $O_0(x_1)$,
    $O_0(x_2)$,
    $O_1(x_1)$,
    $O_1(x_2)$,
    $O'_0(x_1)$,
    $O'_0(x_2)$,
    $O'_1(x_1)$,
    $O'_1(x_2)$,
    {$A_0(x_1,t)$},
    {$A_0(x_2,t)$},
    {$A_1(x_1,t)$},
    {$A_1(x_2,t)$},
  } {
    \node[minimum size=4mm] at (-1,-\i) {\scriptsize \a};
  }
\end{tikzpicture}
\end{subfigure}
\setlength{\abovecaptionskip}{-1.5em}
\setlength{\belowcaptionskip}{-1em}
\caption{Visualization of the target similarity matrix ($K=2$ with two samples in a batch). The white, blue, and lavender blocks denote the excluded, positive, and negative pairs, respectively. The red area contains the newly introduced negative pairs with dissolving transformations.}
\label{fig:heatmap}
\end{wrapfigure}
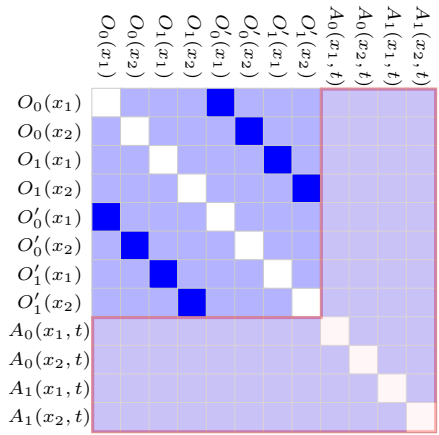
For a single image, we have $3K$ different transformations. With $B$ different images in a batch, yielding $3K\cdot B$ images that are considered jointly. For all possible pairs of images, they can either be a \textit{negative pair}, a \textit{positive pair}, or not be considered in the loss function. We relegate the explanation to an illustration in ~\cref{fig:heatmap}. 
In the top left quadrant of the matrix, we can see the design choices of what constitutes a positive and a negative pair inherited from \cite{tack2020csi}, based on the \textit{NT-Xent} loss \cite{chen2020simple}. The region highlighted in red, is our proposed design for the new \textit{negative pairs} for \textit{dissolving transformations}. The purpose of these newly introduced \textit{negative pairs}  is to learn a representation that can better distinguish between fine-grained semantically meaningful features. The contrastive loss for each image sample can be computed as follows:
\begin{align}
\small
    \ell_{i,j} =-\log
    \frac{
    \exp(\text{sim}({z}_i,{z}_j)/\tau)
    }
    {
    \sum_{k=1}^{3N}\mymathbb{1}_{k,i} \cdot (\exp(\text{sim}({z}_i,{z}_k)) / \tau)
    },
    \quad
    \mymathbb{1}_{k,i} =
    \begin{cases}
        0\quad i=k, \\  
        1\quad otherwise,
    \end{cases}
    \label{eq:xt-xent}
\end{align}
where $N$ is the number of samples (\ie $N=B\cdot K$), $sim(z,\hat{z})=z\cdot \hat{z} / ||z|| ||\hat{z}||$, and $\tau$ is a temperature hyperparameter to control the penalties of negative samples.



As mentioned, the positive pairs are selected from $O_i(\cdot)$ and $O'_j(\cdot)$ branches only when $i=j$. The proposed \textit{feature-amplified NT-Xent} loss can therefore be expressed as:
{\small 
\begin{align}
    \mathcal{L}_{con} =
    & \dfrac{1}{3BK} \dfrac{1}{ |\{x_+\}|} \sum \ell_{i,j} \cdot 
    \begin{cases}
      0 & {\mymathbb{1}_{i,j}\in\{x_-\}}  \\
      1 & {\mymathbb{1}_{i,j}\in\{x_+\}}  
    \end{cases},
    \label{eq:con_loss}
\end{align}}
where $\{x_+\}$ and $\{x_-\}$ denote the positive and negative pairs, and $|\{x_+\}|$ is the number of positive pairs.

Additionally, an auxiliary softmax classifier $f_\theta$ is used to predict which shifting transformation is applied for a given input $x$, resulting in $p_{cls}(y^{S}|x)$. With the union of non-dissolving and dissolving transformed samples $\mathcal{X}_{\mathcal{S}\cup \mathcal{A}}$, the classification loss is defined as:
\begin{equation}
    \small
    \mathcal{L}_{cls}=\frac{1}{3B}\frac{1}{K}\sum_{k=0}^{K-1}\sum_{\hat{x}\in\mathcal{X}_{\mathcal{S}\cup \mathcal{A}}} -\log p_{cls}(y^{S}|\hat{x}).
\end{equation}
The final training loss is hereby defined as:
\begin{equation}
    \small
\mathcal{L}_{DIA}=\mathcal{L}_{con}+\gamma\cdot\mathcal{L}_{cls},
\end{equation}
where $\gamma$ is set to 1 in this work.

\subsection{The Score functions}
\label{sec:score}

During inference, we adopt an anomaly score function that consists of two parts: (1) $s_{con}$ sums the anomaly scores over all shifted transformations, in addition to (2) $s_{cls}$ sums the confidence of the shifting transformation classifier. For the $k^{th}$ shifting transformation, given an input image $x$, training example set $\{x_m\}$, and a feature extractor $c$, we have:
\begin{align}
    \small
    s_{con}(\Tilde{x}, \{\Tilde{x}_m\}) = \max_m\; \text{sim}(c(\Tilde{x}_m),c(\Tilde{x})) \cdot ||c(\Tilde{x})||,
    s_{cls}(\Tilde{x}) = W_k f_\theta (\Tilde{x}), \\
    \text{With} \quad \Tilde{x} = T_k(x) \quad \Tilde{x}_m = T_k(x_m)\nonumber
\end{align}
where $\max_m \text{sim} (c(x_m), c(x))$ computes the \textit{cosine similarity} between $x$ and its nearest training sample in $\{x_m\}$, $f_\theta$ is an auxiliary classifier that aims at determining if $x$ is a shifted example or not, and $W_k$ is the weight vector in the linear layer of $p_{cls}(y^{S}|x)$. In practice, with $M$ training samples, balancing terms $\lambda^S_{con}=M/\sum_m s^S_{con}$ and $\lambda^S_{cls}=M/\sum_m s^S_{cls}$ are applied to scale the scores of each shifting transformation $S$. Those balancing terms slightly improve the detection performances, as reported in~\cite{tack2020csi}. Our final anomaly score is $s_{con}(\Tilde{x},\{\Tilde{x}_m\}) \cdot \lambda_{con}^S+s_{cls}(\Tilde{x})\cdot \lambda_{cls}^S$.


\section{Experiments}

\label{sec:exp}

\begin{table*}[t]
    \centering
    \scriptsize
    \setlength{\tabcolsep}{1pt}
    \renewcommand{\arraystretch}{.7}
    \begin{tabular}{l|c|cc | cccc}
        \toprule
            { Methods}
            & \makecell{\tiny Extra\\\tiny Training\\\tiny Data} 
            & \makecell{\tiny Pnuemonia\\\tiny MNIST} 
            & \makecell{\tiny Breast\\\tiny MNIST} 
            & \makecell{\tiny SARS-\\\tiny COV-2}
            & \makecell{\tiny Kvasir\\\tiny Polyp}
            & \makecell{\tiny Retinal\\\tiny OCT}
            & \makecell{\tiny APTOS\\\tiny 2019} \\
        \midrule
        \midrule
        \multicolumn{3}{l}{\tiny Reconstruction-based Methods} \\
        \midrule
        \hspace{.5em} GANomaly \hfill {\color{darkgray} \scalebox{.7}{(ACCV 18)}} &
           $\times$ &
            0.552{\scalebox{.7}{$\pm$0.01}} &
            0.527{\scalebox{.7}{$\pm$0.01}} &
            0.604{\scalebox{.7}{$\pm$0.00}} &
            0.604{\scalebox{.7}{$\pm$0.00}} &
            0.505{\scalebox{.7}{$\pm$0.00}} &
            0.601{\scalebox{.7}{$\pm$0.01}} \\
        \hspace{.5em} \ddag  UniAD \cite{you_2022_nips} \hfill {\color{darkgray} \scalebox{.7}{(NeurIPS 22)}} & 
           $\times$ &
           0.734{\scalebox{.7}{$\pm$0.02}} & 
           0.624{\scalebox{.7}{$\pm$0.01}} & 
           0.636{\scalebox{.7}{$\pm$0.00}} & 
           0.724{\scalebox{.7}{$\pm$0.03}} & 
           0.921{\scalebox{.7}{$\pm$0.01}} & 
           0.874{\scalebox{.7}{$\pm$0.00}} \\
        \midrule
        \multicolumn{3}{l}{\tiny Normalizing flow-based Methods} \\
        \midrule
        \hspace{.5em} \ddag CFlow~\cite{gudovskiy2022cflow} \hfill {\color{darkgray} \scalebox{.7}{(WACV 22)}} &
            $\times$ & 
            0.537{\scalebox{.7}{$\pm$0.01}} &
            0.647{\scalebox{.7}{$\pm$0.01}} &
            0.622{\scalebox{.7}{$\pm$0.01}} &
            0.852{\scalebox{.7}{$\pm$0.03}} &
            0.712{\scalebox{.7}{$\pm$0.02}} &
            0.452{\scalebox{.7}{$\pm$0.01}} \\
        \hspace{.5em} UFlow~\cite{tailanian2022u} \hfill {\color{darkgray} \scalebox{.7}{ }} &
            $\times$ &
            0.792{\scalebox{.7}{$\pm$0.01}} &
            0.631{\scalebox{.7}{$\pm$0.01}} &
            0.653{\scalebox{.7}{$\pm$0.02}} &
            0.562{\scalebox{.7}{$\pm$0.02}} &
            0.630{\scalebox{.7}{$\pm$0.01}} &
            0.731{\scalebox{.7}{$\pm$0.00}}  \\
        \hspace{.5em} FastFlow~\cite{yu2021fastflow} \hfill {\color{darkgray} \scalebox{.7}{ }} &
            $\times$ &
            0.827{\scalebox{.7}{$\pm$0.02}} &
            0.667{\scalebox{.7}{$\pm$0.01}} &
            0.700{\scalebox{.7}{$\pm$0.01}} &
            0.516{\scalebox{.7}{$\pm$0.03}} &
            0.744{\scalebox{.7}{$\pm$0.01}} &
            0.772{\scalebox{.7}{$\pm$0.02}} \\
        \midrule
        \multicolumn{3}{l}{\tiny Teacher-Student Methods} \\
        \midrule
        \hspace{.5em} KDAD \cite{salehi2021multiresolution} \hfill {\color{darkgray} \hspace{.8cm} \scalebox{.7}{(CVPR 21)}} &
           $\times$ &
            0.378{\scalebox{.7}{$\pm$0.02}} & 
            0.611{\scalebox{.7}{$\pm$0.02}} & 
            0.770{\scalebox{.7}{$\pm$0.01}} & 
            0.775{\scalebox{.7}{$\pm$0.01}} & 
            0.801{\scalebox{.7}{$\pm$0.00}} & 
            0.631{\scalebox{.7}{$\pm$0.01}}  \\
        \hspace{.5em} RD4AD \cite{Deng_2022_CVPR} \hfill{\color{darkgray} \scalebox{.7}{(CVPR 22)}} &
            \checkmark &
            0.815{\scalebox{.7}{$\pm$0.01}} &
            \textbf{0.759}{\scalebox{.7}{$\pm$0.02}} & 
            {0.842}{\scalebox{.7}{$\pm$0.00}} & 
            0.757{\scalebox{.7}{$\pm$0.01}} & 
            \textbf{0.996}{\scalebox{.7}{$\pm$0.00}} & 
            0.921{\scalebox{.7}{$\pm$0.00}} \\
        \hspace{.5em} \dag Transformly \cite{Cohen_2022_CVPR} \hfill {\color{darkgray} \scalebox{.7}{(CVPR 22)}}&
            \checkmark &
            0.821{\scalebox{.7}{$\pm$0.01}} & 
            0.738{\scalebox{.7}{$\pm$0.04}} & 
            0.711{\scalebox{.7}{$\pm$0.00}} & 
            0.568{\scalebox{.7}{$\pm$0.00}} &
            0.824{\scalebox{.7}{$\pm$0.01}} & 
            0.616{\scalebox{.7}{$\pm$0.01}} \\
        \hspace{.5em} \ddag EfficientAD~\cite{batzner2024efficientad} \hfill {\color{darkgray} \scalebox{.7}{(CVPR 24)}} &
            \checkmark &
            0.686{\scalebox{.7}{$\pm$0.02}} &
            0.696{\scalebox{.7}{$\pm$0.03}} &
            0.711{\scalebox{.7}{$\pm$0.02}} &
            0.753{\scalebox{.7}{$\pm$0.03}} &
            0.826{\scalebox{.7}{$\pm$0.02}} &
            0.763{\scalebox{.7}{$\pm$0.02}}\\
        \midrule
        \multicolumn{3}{l}{\tiny Memory Bank-Based Methods} \\
        \midrule
        \hspace{.5em} CFA \hfill {\color{darkgray} \scalebox{.7}{(IEEE Access 22)}} & 
           $\times$ & 
            {0.716}{\scalebox{.7}{$\pm$0.01}} &
            {0.678}{\scalebox{.7}{$\pm$0.02}} &
            {0.424}{\scalebox{.7}{$\pm$0.03}} &
            {0.354}{\scalebox{.7}{$\pm$0.01}} &
            {0.472}{\scalebox{.7}{$\pm$0.01}} &
            {0.796}{\scalebox{.7}{$\pm$0.01}} \\
        \hspace{.5em} PatchCore \hfill {\color{darkgray} \scalebox{.7}{(CVPR 22)}} &
           $\times$ & 
            {0.737}{\scalebox{.7}{$\pm$0.01}} &
            {0.700}{\scalebox{.7}{$\pm$0.02}} &
            {0.654}{\scalebox{.7}{$\pm$0.01}} &
            {0.832}{\scalebox{.7}{$\pm$0.01}} &
            {0.758}{\scalebox{.7}{$\pm$0.01}} &
            {0.583}{\scalebox{.7}{$\pm$0.01}} \\
        \midrule
        \multicolumn{3}{l}{\tiny Contrastive Learning-Based Methods} \\
        \midrule
        \hspace{.5em} Meanshift \cite{reiss2021mean} \hfill {\color{darkgray} \scalebox{.7}{(AAAI 23)}} & 
           $\times$ &
            0.818{\scalebox{.7}{$\pm$0.02}} & 
            0.648{\scalebox{.7}{$\pm$0.01}} & 
            0.767{\scalebox{.7}{$\pm$0.03}} & 
            0.694{\scalebox{.7}{$\pm$0.05}} & 
            0.438{\scalebox{.7}{$\pm$0.01}} & 
            0.826{\scalebox{.7}{$\pm$0.01}} \\
         \hspace{.5em} CSI \cite{tack2020csi}{ \hspace{.2em}\color{darkgray} \tiny Baseline} \hfill {\color{darkgray} \scalebox{.7}{(NeurIPS 20)}} & 
           $\times$ &
            0.834{\scalebox{.7}{$\pm$0.03}} & 
            0.546{\scalebox{.7}{$\pm$0.03}} & 
            0.785{\scalebox{.7}{$\pm$0.02}} & 
            0.609{\scalebox{.7}{$\pm$0.03}} & 
            0.803{\scalebox{.7}{$\pm$0.00}} &
            {0.927}{\scalebox{.7}{$\pm$0.00}} \\
        \midrule
        \midrule
        \hspace{.5em} DIA { \hspace{1.5em}\color{darkgray} \tiny Ours} & 
           $\times$ &
            \textbf{0.903}{\scalebox{.7}{$\pm$0.01}} & 
            {0.750}{\scalebox{.7}{$\pm$0.03}} & 
            \textbf{0.851}{\scalebox{.7}{$\pm$0.03}} & 
            \textbf{0.860}{\scalebox{.7}{$\pm$0.04}} & 
            {0.944}{\scalebox{.7}{$\pm$0.00}} & 
            \textbf{0.934}{\scalebox{.7}{$\pm$0.00}} \\
        \bottomrule
        \multicolumn{8}{l}{}\\
        \multicolumn{8}{l}{\hspace{-1em} \scriptsize \dag Transformaly is trained under unimodel settings as the original paper.}\\
        \multicolumn{8}{l}{\hspace{-1em} \scriptsize \ddag Not support $32\times 32$ resolution, where $128\times 128$ resolution is used for *MNIST datasets.} \\
        \multicolumn{8}{l}{ \scriptsize Only 4500 images of the OCT dataset for \textit{PatchCore} are used due to it is the cap for A100.}\\
    \end{tabular}
    \setlength{\belowcaptionskip}{-1.em}
    \setlength{\abovecaptionskip}{0em}
    \caption{Semi-supervised fine-grained medical anomaly detection results.}
    \label{tab:med}
\end{table*}

\subsection{Experiment Setting}

We evaluated our methods on six datasets with various imaging protocols (\eg CT, OCT, endoscopy, retinal fundus) and areas (\eg chest, breast, colon, eye). In particular, we experiment on low-resolution datasets of \textit{Pnuemonia MNIST} and \textit{Breast MNIST}, and higher resolution datasets of \textit{SARS-COV-2}, \textit{Kvasir-Polyp}, \textit{Retinal-OCT}, and \textit{APTOS-2019}. A detailed description is in \cref{sec:med_data}.

We performed semi-supervised anomaly detection that uses only the normal class for training, namely, the healthy samples. Then we output the anomaly scores for each data instance to evaluate the anomaly detection performance. We use the area under the receiver operating characteristic curve (AUROC) as the metric. All the presented values are computed by averaging at least three runs. Technical details can be found in~\Cref{sec:tech_detail}. Technically, we use ResNet18 as the backbone model and a batch size of 32. We adopted rotation as the \textit{shifting transformations}, with a fixed $K=4$ for $0\degree$, $90\degree$, $180\degree$, $270\degree$.
For the \textit{Kvasir-Polyp} dataset, we used perm (\ie jigsaw transformation) since gastrointestinal images are rotation-invariant (details in \Cref{sec:suppl_shift}).
For \textit{dissolving transformations}, all diffusion models are trained on $32\times 32$ images. The diffusion step $t$ is randomly sampled from $t\sim U(100, 200)$ for \textit{Kvasir-Polyp} and $t\sim U(30, 130)$ for the other datasets.
For high-resolution datasets, we downsampled images to $32\times 32$ for feature dissolving and then resized them back, avoiding massive computations. Results for different dissolving transformation resolutions are in~\Cref{sec:reso_diss}.

\subsection{Results}

We compare against 14 previous methods to showcase the performances of our method. Most selected methods are designed for fine-grained anomaly detection or medical anomaly detection.
As shown in \cref{tab:med}, previous work is underperforming or unstable across various fine-grained anomaly detection datasets. Methods that do not leverage external data generally perform less effectively. In contrast, our approach, which employs a fine-grained feature learning strategy, achieves consistently strong and reliable results across all datasets without relying on pretrained models. This highlights the reliability and effectiveness of our strategy, underscoring its superiority in handling diverse medical data modalities and anomaly patterns with stable performances. Notably,  our method beats all other methods on four out of six datasets. \textit{RD4AD} takes advantage of pretrained models and achieves better performances on two datasets. In addition, we significantly outperform the baseline \textit{CSI} on all datasets, thereby clearly demonstrating the value of our novel fine-grained feature learning paradigm.

\section{Ablation Studies}

\label{sec:ablations}


This section presents a series of ablation studies to understand how our proposed method works under different configurations and parameter settings. 
In addition, we present results with heuristic blurring methods and shifting transformations in~\cref{sec:heuristic}, along with the different designs of similarity matrix and non-medical datasets provided in~\cref{sec:sim_design}.

\subsection{Dissolving Transformation Steps}

We randomly sample dissolving step $t$ from a uniform distribution $U(a,b)$. This experiment investigates various sampling ranges. We establish the minimum step at 30 to ensure minimal changes to the image and assess effectiveness over a 100-step interval. As indicated in~\cref{tab:steps}, lower steps generally yield better results. The lower step dissolves fine-grained features without significantly altering the coarse-grained image appearance. The model can then focus on the dissolved fine-grained features. Kvasir dataset involves polyps as anomalies, which are pronounced (in the pixel space) compared to the anomalies in other datasets. Consequently, a slightly higher $t$ can lead to enhanced performance.

\begin{minipage}[b][][b]{0.48\linewidth}
    \vspace{1em}
    \scriptsize
    \centering
    \begin{tabular}{c|c cccc}
        \toprule
        \shortstack[c]{Step Range}
            & \shortstack[c]{SARS\\COV-2}
            & \shortstack[c]{Kvasir\\Polyp}
            & \shortstack[c]{Retinal\\OCT}
            & \shortstack[c]{APTOS\\2019} \\
        \midrule
        (~~30, 130) 
            & \textbf{0.851}
            & 0.796
            & \textbf{0.919}
            & \textbf{0.934} \\
        (130, 230)  
            & 0.827
            & \textbf{0.860}
            & 0.895
            & 0.920 \\
        (230, 330)    
            & 0.790
            & 0.775
            & 0.908
            & 0.923 \\
        (330, 430)    
            & 0.815
            & 0.763
            & 0.896
            & 0.926 \\
        (430, 530)    
            & 0.803
            & 0.615
            & 0.905
            & 0.926 \\
        \bottomrule
    \end{tabular}
    \captionof{table}{Different diffusion step range.}
    \label{tab:steps}
\end{minipage}\hfill
\begin{minipage}[b][][b]{0.48\linewidth}
    \scriptsize
    \centering
    \begin{tabular}{
        @{\hskip3pt}l@{\hskip3pt}
        |
        @{\hskip3pt}c@{\hskip3pt}
        @{\hskip3pt}c@{\hskip3pt}
    }
        \toprule
        Datasets & \shortstack[c]{DIA\\($\gamma=0.1$)} & \shortstack[c]{DIA\\($\gamma=1$)}  \\
         \midrule
         \shortstack[c]{PneumoniaMNIST} & 0.745 & 0.903 \\
         \shortstack[c]{Kvasir-Polyp} & 0.679 & 0.860 \\
        \bottomrule
    \end{tabular}
    \captionof{table}{Different training data ratios.}
    \label{tab:ratio}
\end{minipage}


\subsection{The Role of Diffusion Models}
\label{sec:role}


Given the challenges of acquiring additional medical data, we evaluate how diffusion models affect anomaly detection performances. Specifically, we limit the training data ratio ($\gamma$) for diffusion models to simulate less optimal diffusion models, while keeping other settings unchanged. 
This experiment examines how anomaly detection performances are impacted when deployed with underperforming diffusion models with insufficient training data.
We evaluate on two small datasets where 5856 images are in \textit{PneumoniaMNIST} and 8000 images are in \textit{Kvasir-Polyp}.
As shown in \cref{tab:ratio}, a significant performance drop happened. Thus, better performance of anomaly detection can be obtained with better-trained diffusion models.

\begin{figure}[h]
\centering
\begin{subfigure}[t]{.172\linewidth}
  \centering
  \includegraphics[width=.98\linewidth,trim={0 9.2cm 4.6cm 0}, clip]{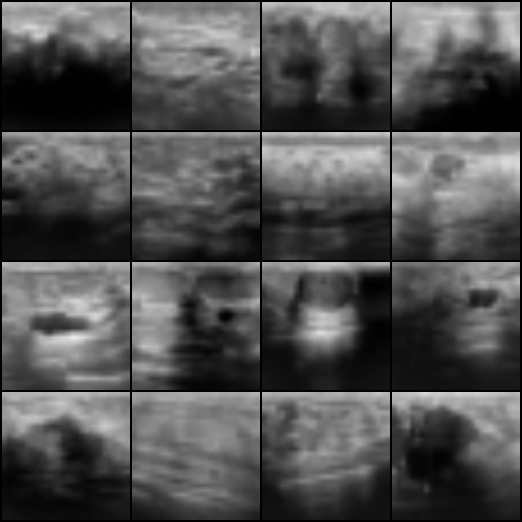}
  \includegraphics[width=.98\linewidth,trim={0 9.2cm 4.6cm 0}, clip]{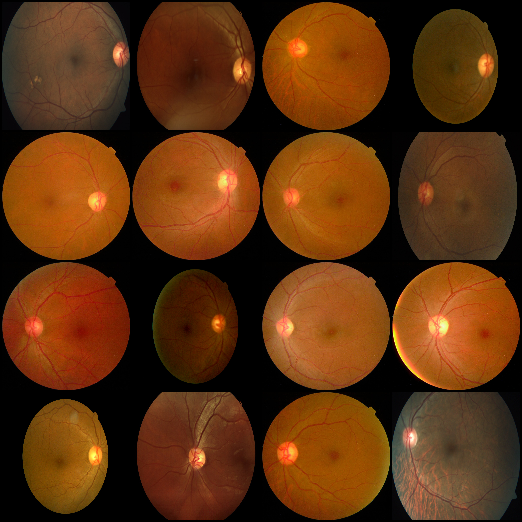}
  \includegraphics[width=.98\linewidth,trim={0 9.2cm 4.6cm 0}, clip]{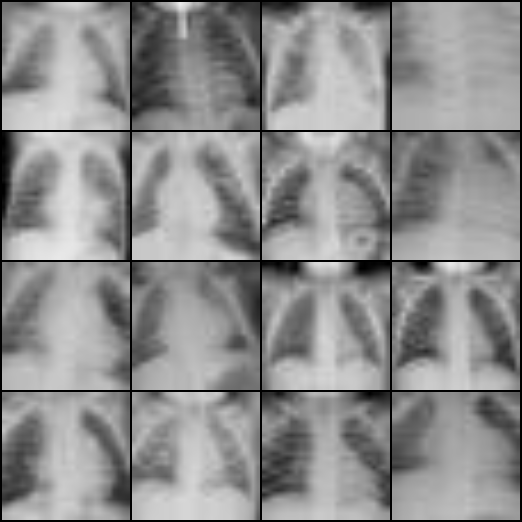}
  \caption{\footnotesize Input}
  \label{fig:wrong_sfig1}
\end{subfigure}%
\hspace{.3em}
\begin{subfigure}[t]{.172\linewidth}
  \centering
  \includegraphics[width=.98\linewidth,trim={0 9.2cm 4.6cm 0}, clip]{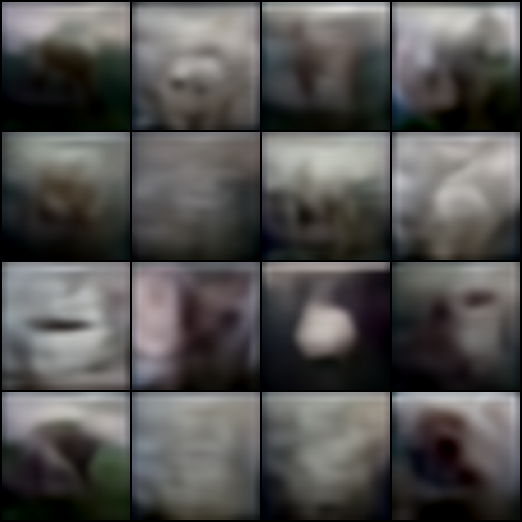}
  \includegraphics[width=.98\linewidth,trim={0 9.2cm 4.6cm 0}, clip]{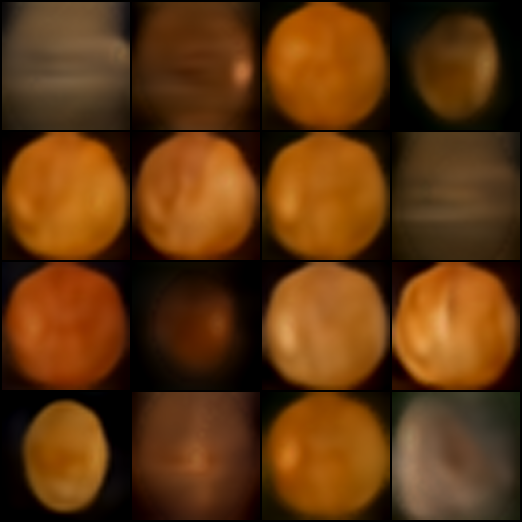}
  \includegraphics[width=.98\linewidth,trim={0 9.2cm 4.6cm 0}, clip]{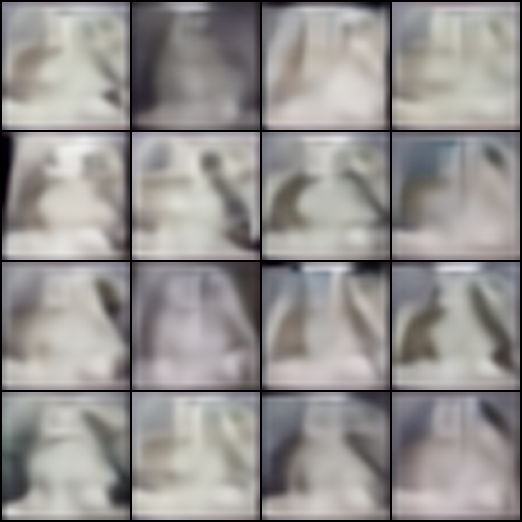}
  \caption{\scriptsize $C,t\!=\!200$}
  \label{fig:wrong_sfig1.5}
\end{subfigure}%
\begin{subfigure}[t]{.172\linewidth}
  \centering
  \includegraphics[width=.98\linewidth,trim={0 9.2cm 4.6cm 0}, clip]{misc/samples_hd/ddmp_p_sample-breastmnist-step_200-x_start.png}
  \includegraphics[width=.98\linewidth,trim={0 9.2cm 4.6cm 0}, clip]{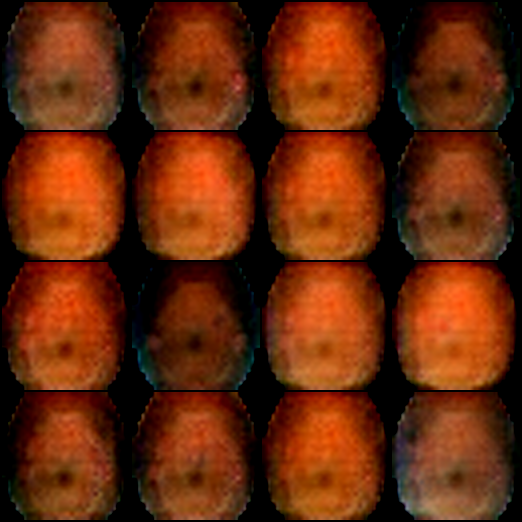}
  \includegraphics[width=.98\linewidth,trim={0 9.2cm 4.6cm 0}, clip]{misc/samples_hd/ddmp_p_sample-pneumoniamnist-step_200-x_start.png}
  \caption{\scriptsize $M,t\!=\!200$}
  \label{fig:wrong_sfig2}
\end{subfigure}%
\hspace{.3em}
\begin{subfigure}[t]{.172\linewidth}
  \centering
  \includegraphics[width=.98\linewidth,trim={0 9.2cm 4.6cm 0}, clip]{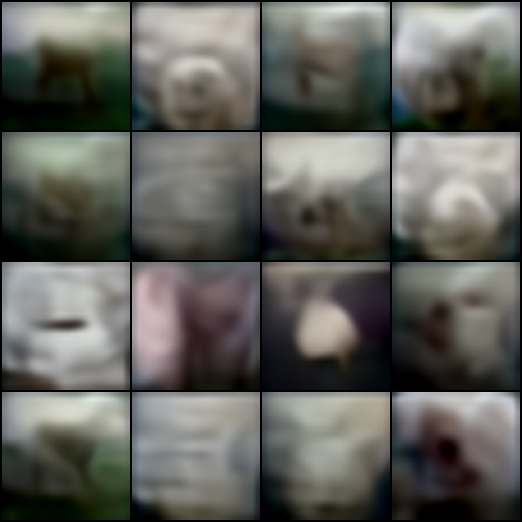}
  \includegraphics[width=.98\linewidth,trim={0 9.2cm 4.6cm 0}, clip]{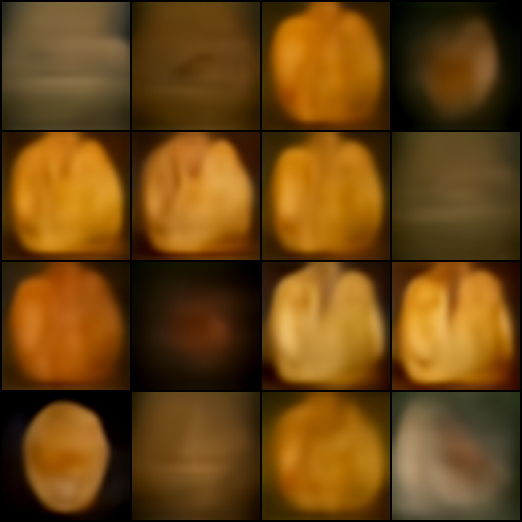}
  \includegraphics[width=.98\linewidth,trim={0 9.2cm 4.6cm 0}, clip]{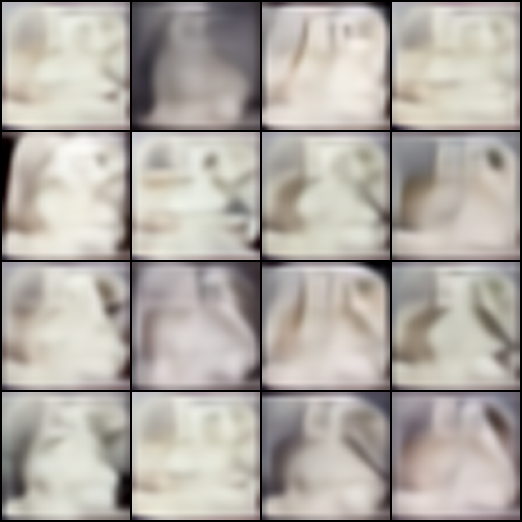}
  \caption{\scriptsize $C,t\!=\!400$}
  \label{fig:wrong_sfig3}
\end{subfigure}%
\begin{subfigure}[t]{.172\linewidth}
  \centering
  \includegraphics[width=.98\linewidth,trim={0 9.2cm 4.6cm 0}, clip]{misc/samples_hd/ddmp_p_sample-breastmnist-step_400-x_start.png}
  \includegraphics[width=.98\linewidth,trim={0 9.2cm 4.6cm 0}, clip]{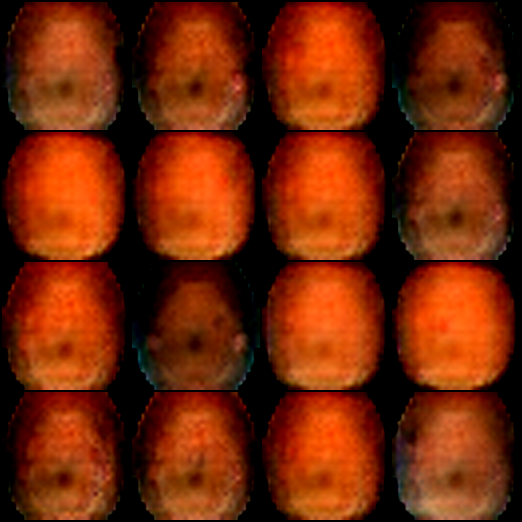}
  \includegraphics[width=.98\linewidth,trim={0 9.2cm 4.6cm 0}, clip]{misc/samples_hd/ddmp_p_sample-pneumoniamnist-step_400-x_start.png}
  \caption{\scriptsize $M,t\!=\!400$}
  \label{fig:wrong_sfig4}
\end{subfigure}%
\setlength{\belowcaptionskip}{-2.em}
\setlength{\abovecaptionskip}{0em}
\caption{Dissolving Transformations using different diffusion models. $C$ and $M$ denote if the dissolving transformation is performed based on the diffusion models trained on CIFAR10 or the corresponding dataset, respectively.}
\label{fig:false_dissolving}
\end{figure}

A natural next question is, can one utilize well-trained diffusion models to perform dissolving transformations on non-training domains? A well-trained diffusion model is attuned to the attributes of its training dataset. Consequently, it may incorrectly dissolve features if the presented image deviates from the training set. \Cref{fig:false_dissolving} presents the different dissolving effects using diffusion models trained on different datasets. The visual evidence suggests that a data-specific diffusion model accurately dissolves the correct instance-specific features and attempts to revert images towards a more generalized form characteristic of the dataset. In contrast, a diffusion model trained on the CIFAR dataset tends to dissolve the image in a chaotic manner, failing to maintain the image's inherent shape. Additional demonstration with stable diffusion is in~\Cref{sec:dslv-sd}.

\subsection{Rotate vs. Perm}
Rotate and perm (\ie jigsaw transformation) are reported as the most performant \textit{shifting transformations}~\cite{tack2020csi}. This experiment evaluates their performances under fine-grained settings. As shown in \cref{tab:transform}, the rotation transformation outperforms the perm transformation for most datasets. Perm transformation performs better on the Kvasir dataset since the endoscopic images can be rotation-invariant. In general, the selection of shifting transformations should ease the categorization difficulties associated with the correct shifting distributions. Additional results are in~\cref{sec:suppl_shift}.

\begin{table}[h]
    \scriptsize
    \centering
    \setlength{\tabcolsep}{3pt}
    \begin{tabular}{c|c c c c}
        \toprule
         Method & SARS-COV-2 & Kvasir-Polyp & Retinal-OCT & APTOS-2019  \\
         \midrule
         DIA-Perm &
            0.841{\scriptsize$\pm$0.01} &
            \textbf{0.860}{\scriptsize$\pm$0.01} &
            0.890{\scriptsize$\pm$0.02} &
            0.926{\scriptsize$\pm$0.00}
            \\
         DIA-Rotate & 
            \textbf{0.851}{\scriptsize$\pm$0.03} &
            {0.813}{\scriptsize$\pm$0.03} &
            \textbf{0.944}{\scriptsize$\pm$0.01} &
            \textbf{0.934}{\scriptsize$\pm$0.00}
            \\
        \bottomrule
    \end{tabular}
    \setlength{\belowcaptionskip}{-2.5em}
    \setlength{\abovecaptionskip}{.5em}
    \caption{Using rotate or perm for shifting transformation.}
    \label{tab:transform}
\end{table}

\subsection{The Resolution of Feature Dissolved Samples}
\label{sec:reso_diss}

We use \textit{feature-dissolved} samples with a resolution of 32$\times$32, which significantly improves the anomaly detection performances. Notably, the downsample-upsample routine also dissolves fine-grained features. This experiment investigates the effects of different resolutions for feature-dissolved samples. The experiments adopt 256, 128, 32 batchsizes for the resolution of $32\times32$, $64\times64$, $128\times128$, respectively.  As shown in \cref{tab:resolution} and \cref{tab:complexity}, the computational cost increases dramatically with increased resolutions, while it can hardly boost model performances.

The variations in performance across different resolutions are attributed to two main factors. Firstly, the size of training samples impacts this. In larger datasets such as APTOS and Retinal-OCT, the performance degradation is less pronounced. This is because higher-resolution diffusion models require more training data. Secondly, the nature of discriminative features plays a role. High-resolution images naturally contain more details. In datasets like APTOS, where disease indicators are subtler in pixel space (\eg hemorrhages or thinner blood vessels), the performance drop is minimal. In fact, 64x64 resolution images even outperform 32x32 ones for APTOS. Conversely, in datasets like Retinal-OCT, where crucial features are more prominent in pixel space (\eg edemas), lower-resolution images help the model concentrate on these more apparent features. Notably, the computational cost of higher-resolution dissolving transformations is dramatically increased. Our results indicate that a resolution of 32x32 strikes an optimal performance for dissolving effects and computational efficiency.

\begin{minipage}[b][][b]{0.46\linewidth}
    \scriptsize
    \centering
    \setlength{\tabcolsep}{3pt}
    \begin{tabular}{c|c c c c}
        \toprule
        \shortstack[c]{Dslv.\\Size}
            & \shortstack[c]{SARS\\COV-2}
            & \shortstack[c]{Kvasir\\Polyp}
            & \shortstack[c]{Retinal\\OCT}
            & \shortstack[c]{APTOS\\2019}  \\
         \midrule
         32 &
            \textbf{0.851} & 
            \textbf{0.860} & 
            \textbf{0.944} & 
            0.934 
            \\
         64 & 
            0.803 & 
            0.721 & 
            0.922 & 
            \textbf{0.937} 
            \\
         128 & 
            0.807 & 
            0.730 & 
            0.930 & 
            0.905 
            \\
        \bottomrule
    \end{tabular}
    \setlength{\belowcaptionskip}{.8em}
    \captionof{table}{Different resolutions for dissolving transformations.}
    \label{tab:resolution}
\end{minipage}
\begin{minipage}[b][][b]{0.48\linewidth}

    \centering
    \scriptsize
    \begin{tabular*}{1.\columnwidth}{@{\extracolsep{\fill}}ccccc@{}}
        \toprule
        Res. 
            & w/o 
            & 32$\times$32 
            & 64$\times$64 
            & 128$\times$128 \\
        \midrule
        Params (M) & 11.2 & 19.93 & 19.93 & 19.93 \\
        MACs \space (G) & 1.82 & 2.33 & 3.84 & 9.90 \\
        \bottomrule
    \end{tabular*}
    \captionof{table}{Multiply–accumulate operations (MACs) for different resolutions of dissolving transformations. \textit{w/o} denotes no dissolving transformation applied.}
    \label{tab:complexity}
\end{minipage}

\section{Discussion}
\label{sec:discussion}

Diffusion models work by gradually adding noise to an image over several steps, and then a UNet is employed to learn to reverse this process.
During the training of diffusion models, the UNet learns to predict the noise that was added at each step of the diffusion process. This process indirectly teaches the UNet about the underlying structure and characteristics of the data in the dataset.
Essentially, the proposed \textit{dissolving transformation} executes a standalone reverse diffusion to reverse the "noise" on non-noisy input images directly.
Notably, it still operates under the assumption that there is noise to be removed. Consequently, it interprets the instance-specific fine details and textures of the non-noisy image as noise and attempts to remove them (which we refer to as "dissolve" in our context), as illustrated in~\cref{fig:demo}.
With non-noisy input images from a non-training domain, the diffusion model fails to interpret the correct instance-specific fine details and, therefore, fails to remove the correct features inside the image, as illustrated in~\cref{fig:false_dissolving}. We show additional qualitative results in~\Cref{sec:dslv-sd}.

Medical image data is particularly suitable for the proposed \textit{dissolving transformations}.
Different from other data domains, medical images typically feature a consistent prior, commonly referred to as "atlas" in the medical domain, which is an average representation of a specific patient population, onto which more detailed, instance-specific (discriminative) features are superimposed.
For instance, chest X-ray images generally have a gray chest shape on a black background, with additional instance-specific features like bones, tumors, or other pathological findings, being superimposed on top.
Those instance-specific features are interpreted by the UNet as "noise" and then removed by the reverse diffusion process.
By tuning the hyperparameter $t$, this process allows for the gradual removal of the most instance-specific features, moving towards the atlas representation of the given image.
The feature-dissolved atlas representation serves as a reference for comparison to identify clinically significant changes, while the removed features typically contain pivotal pathological findings.
Therefore, to amplify these removed critical features, we deploy a contrastive learning scheme to contrast a given input image and its feature-dissolved counterpart.

\section{Conclusion}

We proposed an intuitive \textit{dissolving is amplifying} (DIA) method to support fine-grained discriminative feature learning for medical anomaly detection. Specifically, we introduced \textit{dissolving transformations} that can be achieved with a pre-trained diffusion model. We use contrastive learning to enhance the difference between images that have been transformed by dissolving transformations and images that have not.  Experiments show \textit{DIA} significantly boosts performance on fine-grained medical anomaly detection without prior knowledge of anomalous features.
One limitation is that our method requires training on diffusion models for each of the datasets. In future work, we would like to extend our method to enhance supervised contrastive learning and fine-grained classification by leveraging the fine-grained feature learning strategy.
\bibliographystyle{splncs04}
\bibliography{main}

\clearpage
\appendix
\setcounter{page}{1}

{
\centering
\Large
\textbf{\thetitle}\\
\vspace{0.5em}Supplementary Material \\
\vspace{1.0em}
}

\section{Settings}
\subsection{Technical Details}
\label{sec:tech_detail}

Our experiments are carried out on the NVIDIA A100 GPU server with CUDA 11.3 and PyTorch 1.11.0. We use a popular diffusion model implementation\footnote{https://github.com/lucidrains/denoising-diffusion-pytorch} to train diffusion models for \textit{dissolving transformation}, and the codebase for DIA is based on the official CSI~\cite{tack2020csi} implementation\footnote{https://github.com/alinlab/CSI}. Additionally, we use the official implementation for all benchmark models included in the paper.

\noindent\textbf{The Training of Diffusion Models.} The diffusion models are trained with a 0.00008 learning rate, 2 step gradient accumulation, 0.995 exponential moving average decay for 25,000 steps. Adam~\cite{kingma2014adam} optimizer and L1 loss are used for optimizing the diffusion model weights, and random horizontal flip is the only augmentation used. Notably, we found that automatic mixed precision~\cite{micikevicius2017mixed} cannot be used for training as it impedes the model from convergence. Commonly, the models trained for around 12,500 steps are already usable for dissolving features and training DIA.

\noindent\textbf{The Training of DIA.} The DIA models are trained with a 0.001 learning rate with cosine annealing~\cite{loshchilov2016sgdr} scheduler, and LARS~\cite{you2017large} optimizer is adopted for optimizing the DIA model parameters. After sampling positive and negative samples, dissolving transformation applies then we perform data augmentation from SimCLR~\cite{chen2020simple}. We randomly select 200 samples from the dataset for training each epoch and we commonly obtain the best model within 200 epochs.

\subsection{Datasets}
\label{sec:med_data}

We evaluated on \textit{MedMNIST} datasets~\cite{medmnistv2}, with image sizes of $28\times 28$:

\vspace{-1em}
\begin{itemize}[leftmargin=*]
    \item \textbf{PneumoniaMNIST} \cite{medmnistv2} consists of 5,856 pediatric chest X-Ray images (pneumonia vs. normal), with a ratio of 9 : 1 for training and validation set.
    \item \textbf{BreastMNIST} \cite{medmnistv2} consists 780 breast ultrasound images (normal and benign tumor vs. malignant tumor), with a ratio of 7 : 1 : 2 for train, validation and test set. 
\end{itemize}
\vspace{-.2em}
We also evaluated multiple high-resolution datasets that are resized to $224\times 224$:
\vspace{-.8em}
\begin{itemize}[leftmargin=*]
    \item \textbf{SARS-COV-2} \cite{Angelov2020} contains 1,252 CT scans that are positive for SARS-CoV-2 infection (COVID-19) and 1,230 CT scans for patients non-infected by SARS-CoV-2.
    \item \textbf{Kvasir-Polyp} \cite{Pogorelov:2017:kvasir} consists the 8,000 endoscopic images, with a ratio of 7 : 3 for training and testing. We remapped the labels to polyp and non-polyp classes.
    \item \textbf{Retinal OCT} \cite{C_Basilan2023-sb} consists 83,484 retinal optical coherence tomography (OCT) images for training, and 968 scans for testing. We remapped the diseased categories (\ie CNV, DME, drusen) to the anomaly class.
    \item \textbf{APTOS-2019} \cite{aptos2019} consists 3,662 fundus images to measure the severity of diabetic retinopathy (DR), with a ratio of 7 : 3 for training and testing. We remapped the four categories (\ie normal, mild DR, moderate DR, severe DR, proliferative DR) to normal and DR classes.
\end{itemize}
\section{Heuristic Alternatives To Dissolving Transformations}
\label{sec:heuristic}

With the proposed \textit{dissolving transformations}, the instance-level features can hereby be emphasized and further focused.
Essentially, \textit{dissolving transformations} use diffusion models to wipe away the discriminative instance features.
In this section, we evaluate our method with naïve alternatives to dissolving transformations, namely, Gaussian blur and median blur.

\begin{figure*}[h]
\centering
\begin{subfigure}{.3\linewidth}
  \captionsetup{font=scriptsize,labelfont=scriptsize}
  \centering
  \includegraphics[width=.98\linewidth,trim={0 13.2cm 7.2cm 0}, clip]{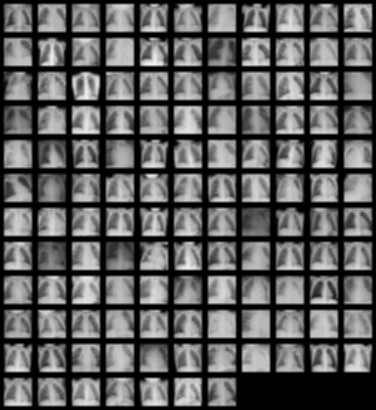}
  \includegraphics[width=.98\linewidth,trim={0 13.2cm 7.2cm 0}, clip]{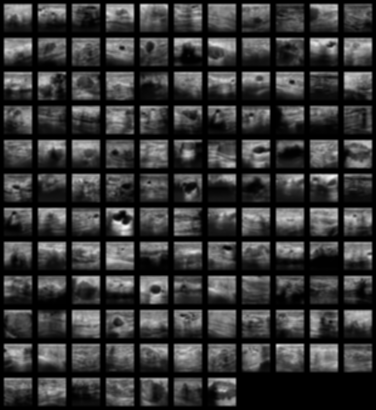}
  \includegraphics[width=.98\linewidth,trim={0 13.2cm 7.2cm 0}, clip]{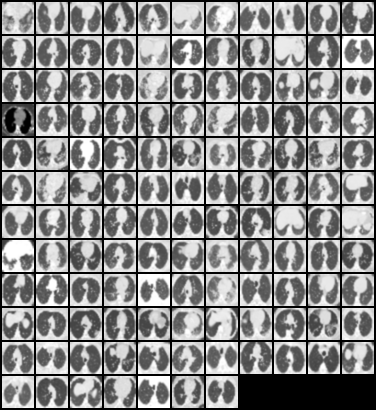}
  \includegraphics[width=.98\linewidth,trim={0 13.2cm 7.2cm 0}, clip]{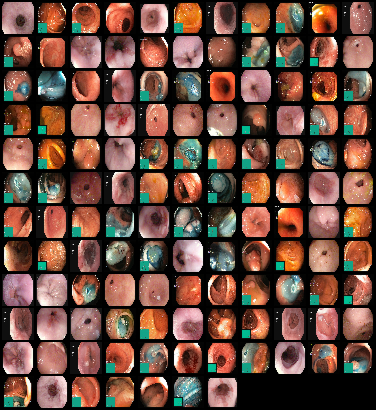}
  \caption{Gaussian ($k$=3)}
  \label{fig:sfig3f_a}
\end{subfigure}
\begin{subfigure}{.3\linewidth}
  \captionsetup{font=scriptsize,labelfont=scriptsize}
  \centering
  \includegraphics[width=.98\linewidth,trim={0 13.2cm 7.2cm 0}, clip]{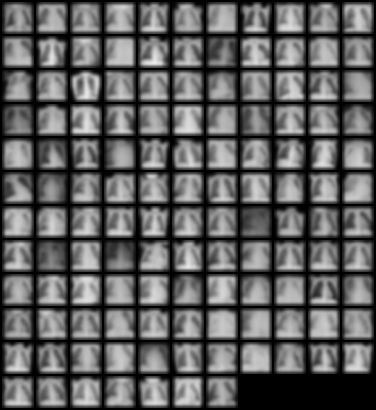}
  \includegraphics[width=.98\linewidth,trim={0 13.2cm 7.2cm 0}, clip]{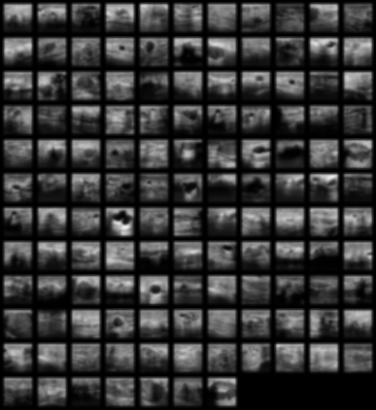}
  \includegraphics[width=.98\linewidth,trim={0 13.2cm 7.2cm 0}, clip]{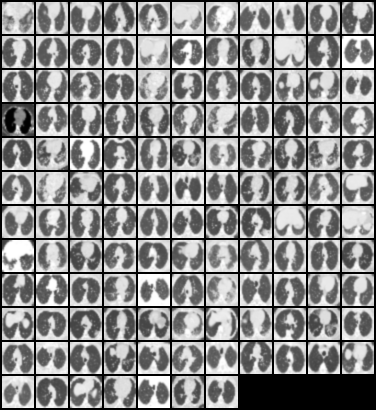}
  \includegraphics[width=.98\linewidth,trim={0 13.2cm 7.2cm 0}, clip]{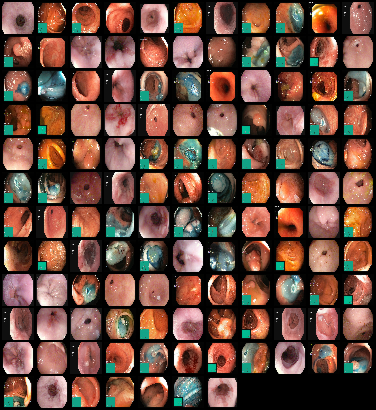}
  \caption{Gaussian ($k$=7)}
  \label{fig:sfig3f_c}
\end{subfigure}
\begin{subfigure}{.3\linewidth}
  \captionsetup{font=scriptsize,labelfont=scriptsize}
  \centering
  \includegraphics[width=.98\linewidth,trim={0 13.2cm 7.2cm 0}, clip]{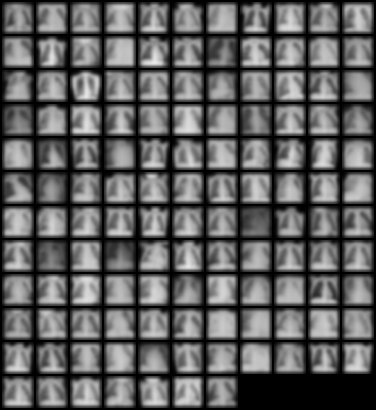}
  \includegraphics[width=.98\linewidth,trim={0 13.2cm 7.2cm 0}, clip]{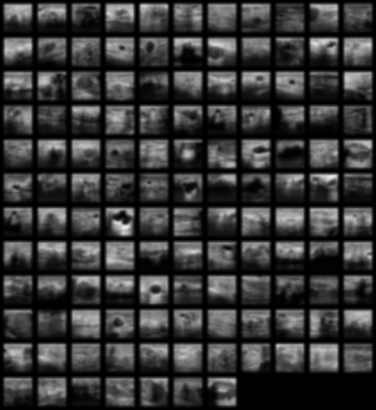}
  \includegraphics[width=.98\linewidth,trim={0 13.2cm 7.2cm 0}, clip]{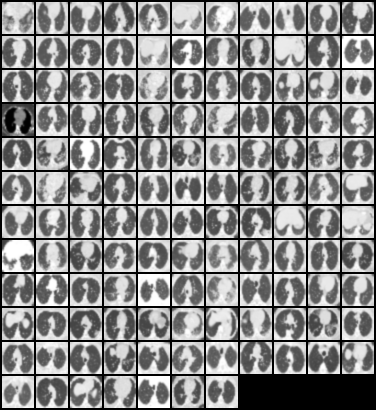}
  \includegraphics[width=.98\linewidth,trim={0 13.2cm 7.2cm 0}, clip]{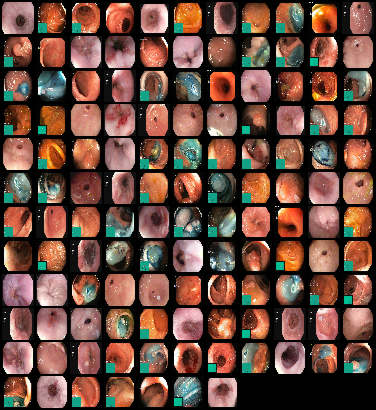}
  \caption{Gaussian ($k$=11)}
  \label{fig:sfig3f_e}
\end{subfigure}

\begin{subfigure}{.3\linewidth}
  \captionsetup{font=scriptsize,labelfont=scriptsize}
  \centering
  \includegraphics[width=.98\linewidth,trim={0 13.2cm 7.2cm 0}, clip]{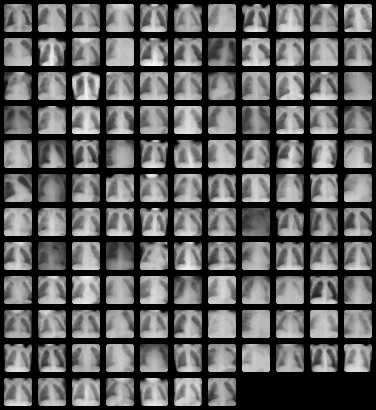}
  \includegraphics[width=.98\linewidth,trim={0 13.2cm 7.2cm 0}, clip]{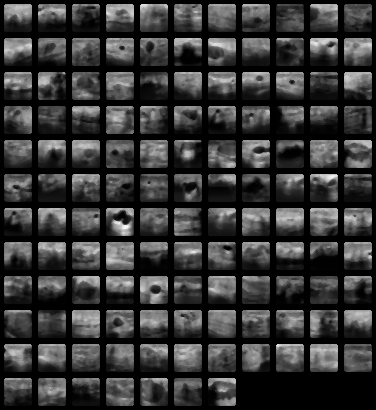}
  \includegraphics[width=.98\linewidth,trim={0 13.2cm 7.2cm 0}, clip]{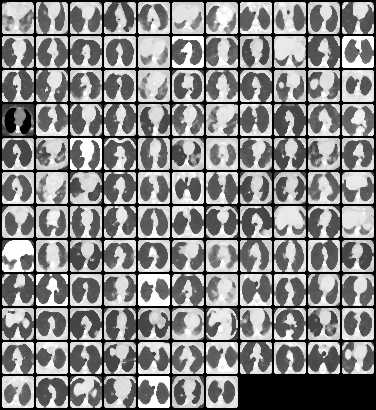}
  \includegraphics[width=.98\linewidth,trim={0 13.2cm 7.2cm 0}, clip]{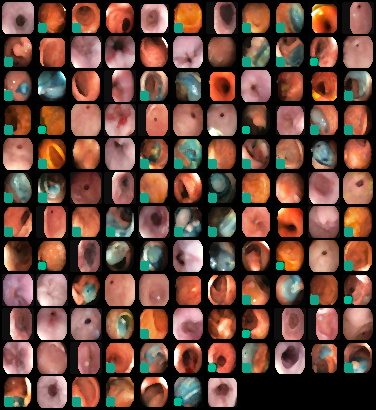}
  \caption{Median ($k$=3)}
  \label{fig:sfig4f_b}
\end{subfigure}
\begin{subfigure}{.3\linewidth}
  \captionsetup{font=scriptsize,labelfont=scriptsize}
  \centering
  \includegraphics[width=.98\linewidth,trim={0 13.2cm 7.2cm 0}, clip]{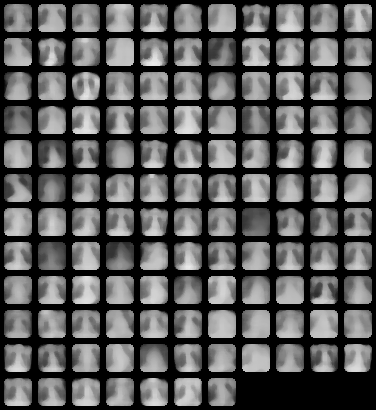}
  \includegraphics[width=.98\linewidth,trim={0 13.2cm 7.2cm 0}, clip]{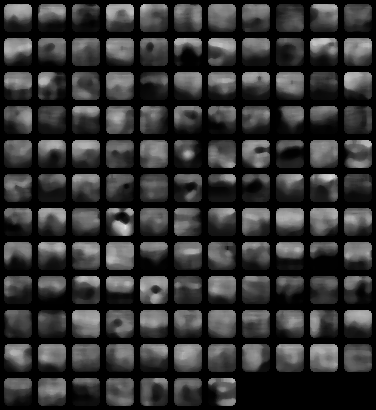}
  \includegraphics[width=.98\linewidth,trim={0 13.2cm 7.2cm 0}, clip]{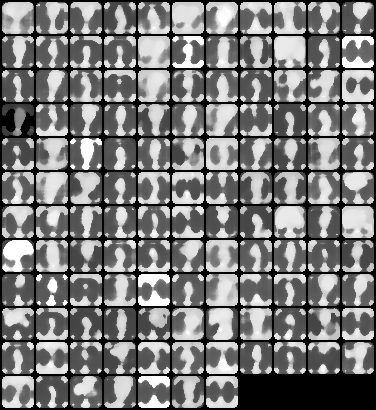}
  \includegraphics[width=.98\linewidth,trim={0 13.2cm 7.2cm 0}, clip]{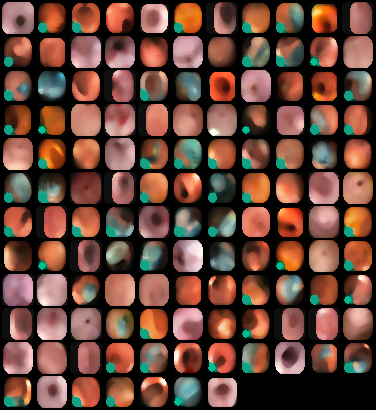}
  \caption{Median ($k$=7)}
  \label{fig:sfig4f_d}
\end{subfigure}
\begin{subfigure}{.3\linewidth}
  \captionsetup{font=scriptsize,labelfont=scriptsize}
  \centering
  \includegraphics[width=.98\linewidth,trim={0 13.2cm 7.2cm 0}, clip]{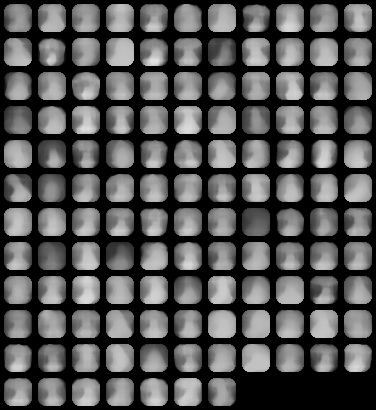}
  \includegraphics[width=.98\linewidth,trim={0 13.2cm 7.2cm 0}, clip]{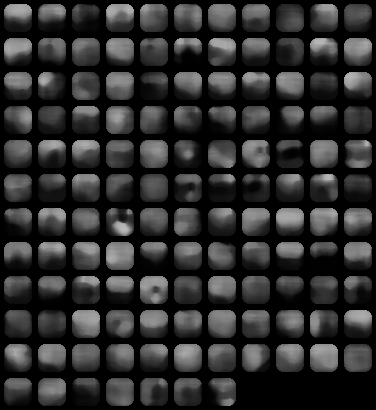}
  \includegraphics[width=.98\linewidth,trim={0 13.2cm 7.2cm 0}, clip]{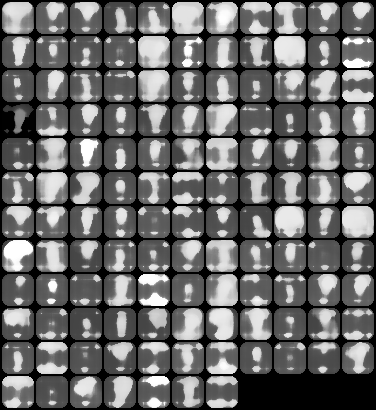}
  \includegraphics[width=.98\linewidth,trim={0 13.2cm 7.2cm 0}, clip]{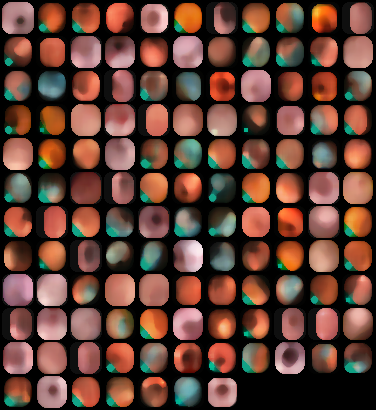}
  \caption{Median ($k$=11)}
  \label{fig:sfig4f_f}
\end{subfigure}%
\caption{Heuristic alternatives to dissolving transformations with various kernel sizes. Compared with median blur, Gaussian blur preserves more image semantics.}
\label{fig:demo_filters}
\end{figure*}

\subsection{Different Kernel Sizes}
\label{sec:filter}

We evaluate different kernel sizes for each operation. A visual comparison of those methods is provided in \cref{fig:demo_filters}.
To be consistent with the diffusion feature dissolving process, the same downsampling and upsampling processes are performed for \textit{DIA-Gaussian} and \textit{DIA-Median}. 
Referring to \cref{tab:med}, though less performant, the heuristic image filtering operations can also contribute to the fine-grained anomaly detection tasks with a significant performance boost against the baseline CSI method.  

\begin{table}[h]
    \scriptsize
    \centering
    \captionsetup[table]{hypcap=false}
    \begin{tabular}{c|ccccc}
        \toprule
        Dataset & kernel size & DIA-Gaussian & DIA-Median \\
        \midrule
        \multirow{3}{*}{\shortstack[c]{pneumonia \\ MNIST}}
            & 3  & 0.845{\scriptsize$\pm$0.01} & 0.779{\scriptsize$\pm$0.03} \\
            & 7  & 0.839{\scriptsize$\pm$0.04} & \textbf{\color{azure}0.872}{\scriptsize$\pm$0.01}  \\
            & 11 & \textbf{\color{azure}0.856}{\scriptsize$\pm$0.02} & 0.678{\scriptsize$\pm$0.07}  \\
        \midrule
        \multirow{3}{*}{\shortstack[c]{breast \\ MNIST}}
            & 3  & 0.541{\scriptsize$\pm$0.01} & 0.641{\scriptsize$\pm$0.03}  \\
            & 7  & 0.653{\scriptsize$\pm$0.03} & \textbf{\color{azure}0.689}{\scriptsize$\pm$0.01}  \\
            & 11 & \textbf{\color{azure}0.749}{\scriptsize$\pm$0.05} & 0.542{\scriptsize$\pm$0.04} \\
        \midrule
        \multirow{3}{*}{\shortstack[c]{SARS-\\COV-2}}
            & 3  & 0.813{\scriptsize$\pm$0.02} & \textbf{\color{azure}0.837}{\scriptsize$\pm$0.07}  \\
            & 7  & \textbf{\color{azure}0.847}{\scriptsize$\pm$0.00} & 0.809{\scriptsize$\pm$0.03}  \\
            & 11 & 0.802{\scriptsize$\pm$0.01} & 0.793{\scriptsize$\pm$0.02} \\
        \midrule
        \multirow{3}{*}{\shortstack[c]{Kvasir \\ Polyp}}
            & 3  & \textbf{\color{azure}0.629}{\scriptsize$\pm$0.03} & \textbf{\color{azure}0.526}{\scriptsize$\pm$0.02}  \\
            & 7  & 0.586{\scriptsize$\pm$0.02} & 0.514{\scriptsize$\pm$0.05}  \\
            & 11 & 0.579{\scriptsize$\pm$0.01} & 0.495{\scriptsize$\pm$0.04}  \\
        \bottomrule
    \end{tabular}
    \setlength{\abovecaptionskip}{.5em}
    \caption{Heuristic alternatives to dissolving transformations with various kernel sizes. The blue color denotes a suboptimal performance against our proposed dissolving transformations.}
    \label{tab:filter}
\end{table}


Compared against the \textit{dissolving transformations}, those non-parametric heuristic methods dissolve image features regardless of the generic image semantics, resulting in lower performances.
In a way, \textit{dissolving transformations} dissolve instance-level image features with an awareness of discriminative instance features, by learning from the dataset. We therefore believe that the \textit{diffusion models} can serve as a better dissolving transformation method for fine-grained feature learning.

\subsection{Rotate vs. Perm}
\label{sec:suppl_shift}

We supplement \cref{tab:transform} with the heuristic alternatives to dissolving transformations in this section. As shown in \cref{tab:transform_blur},
similar to dissolving transformations, the rotation transformation mostly outperforms the perm transformation.

\begin{table}[H]
    \scriptsize
    \centering
    \setlength\tabcolsep{5pt}
    \begin{tabular}{c|c | c ccc}
        \toprule
        Dataset & transform & Resize Only & Gaussian & Median & Diffusion \\
        \midrule
        \multirow{2}{*}{\shortstack[c]{SARS-\\COV-2}}
            & Perm  & 
                0.768{\scriptsize$\pm$0.01} & 
                0.788{\scriptsize$\pm$0.01} & 
                0.826{\scriptsize$\pm$0.00} & 
                0.841{\scriptsize$\pm$0.01}  \\
            & Rotate  & 
                0.779{\scriptsize$\pm$0.01} & 
                \textbf{0.847}{\scriptsize$\pm$0.00} & 
                \textbf{0.837}{\scriptsize$\pm$0.07} & 
                \textbf{0.851}{\scriptsize$\pm$0.03} \\
        \midrule
        \multirow{2}{*}{\shortstack[c]{Kvasir\\Polyp}}
            & Perm & 
                0.826{\scriptsize$\pm$0.01} & 
                0.712{\scriptsize$\pm$0.02} & 
                0.663{\scriptsize$\pm$0.02} & 
                \textbf{0.860}{\scriptsize$\pm$0.01}  \\
            & Rotate  & 
                0.748{\scriptsize$\pm$0.02} & 
                \textbf{0.739}{\scriptsize$\pm$0.00} & 
                \textbf{0.687}{\scriptsize$\pm$0.01} & 
                {0.813}{\scriptsize$\pm$0.03} \\
        \midrule
        \multirow{2}{*}{\shortstack[c]{Retinal\\OCT}}
            & Perm & 
                0.892{\scriptsize$\pm$0.01} & 
                0.754{\scriptsize$\pm$0.01} & 
                0.747{\scriptsize$\pm$0.03} & 
                0.890{\scriptsize$\pm$0.02}  \\
            & Rotate  & 
                0.873{\scriptsize$\pm$0.01} & 
                \textbf{0.895}{\scriptsize$\pm$0.01} & 
                \textbf{0.876}{\scriptsize$\pm$0.02} & 
                \textbf{0.944}{\scriptsize$\pm$0.01} \\
        \midrule
        \multirow{2}{*}{\shortstack[c]{APTOS\\2019}}
            & Perm & 
                0.924{\scriptsize$\pm$0.01} & 
                \textbf{0.942}{\scriptsize$\pm$0.00} & 
                \textbf{0.929}{\scriptsize$\pm$0.00} & 
                0.926{\scriptsize$\pm$0.00}  \\
            & Rotate  & 
                0.918{\scriptsize$\pm$0.01} & 
                0.922{\scriptsize$\pm$0.00} & 
                0.918{\scriptsize$\pm$0.00} & 
                \textbf{0.934}{\scriptsize$\pm$0.00} \\
        \bottomrule
    \end{tabular}
    \setlength{\abovecaptionskip}{.5em}
    \caption{Comparison between rotate and perm as shifting transformation.}
    \label{tab:transform_blur}
\end{table}


\subsection{The Resolution of Feature Dissolved Samples}

We supplement \cref{tab:resolution} with heuristic alternatives to dissolving transformations in this section. As shown in \cref{tab:resolution_blur}, those heuristic alternatives are not as performant as the proposed diffusion transformation.

\begin{table}[h]
    \vspace{-1em}
    \scriptsize
    \centering
    \begin{tabular}{c|c cccc}
        \toprule
        Dataset & size & DIA-Gaussian & DIA-Median & DIA-Diffusion \\
        \midrule
        \multirow{3}{*}{\shortstack[c]{SARS-\\COV-2}}
            & 32  & 
                0.847{\scriptsize$\pm$0.00} & 
                0.837{\scriptsize$\pm$0.07} & 
                \textbf{0.851}{\scriptsize$\pm$0.03} \\
            & 64  &
                0.821{\scriptsize$\pm$0.01} & 
                0.839{\scriptsize$\pm$0.01} & 
                0.803{\scriptsize$\pm$0.01} \\
            & 128 &
                0.838{\scriptsize$\pm$0.00} & 
                0.848{\scriptsize$\pm$0.00} &
                0.807{\scriptsize$\pm$0.02} \\
        \midrule
        \multirow{3}{*}{\shortstack[c]{Kvasir \\ Polyp}}
            & 32  & 
                0.629{\scriptsize$\pm$0.03} & 
                0.526{\scriptsize$\pm$0.02} & 
                \textbf{0.860}{\scriptsize$\pm$0.04} \\
            & 64  &
                0.686{\scriptsize$\pm$0.00} & 
                0.575{\scriptsize$\pm$0.02} & 
                0.721{\scriptsize$\pm$0.01} \\
            & 128 &
                0.581{\scriptsize$\pm$0.01} & 
                0.564{\scriptsize$\pm$0.02} &
                0.730{\scriptsize$\pm$0.02} \\
        \midrule
        \multirow{3}{*}{\shortstack[c]{Retinal \\ OCT}}
            & 32  & 
                0.895{\scriptsize$\pm$0.01} & 
                0.876{\scriptsize$\pm$0.02} & 
                \textbf{0.944}{\scriptsize$\pm$0.01} \\
            & 64  &
                0.894{\scriptsize$\pm$0.00} & 
                0.887{\scriptsize$\pm$0.00} & 
                0.922{\scriptsize$\pm$0.00} \\
            & 128 &
                0.908{\scriptsize$\pm$0.01} & 
                0.906{\scriptsize$\pm$0.00} &
                0.930{\scriptsize$\pm$0.00} \\
        \midrule
        \multirow{3}{*}{\shortstack[c]{APTOS \\ 2019}}
            & 32  & 
                0.922{\scriptsize$\pm$0.00} & 
                0.918{\scriptsize$\pm$0.00} & 
                0.934{\scriptsize$\pm$0.00} \\
            & 64  &
                0.910{\scriptsize$\pm$0.00} & 
                0.917{\scriptsize$\pm$0.00} & 
                \textbf{0.937}{\scriptsize$\pm$0.00} \\
            & 128 &
                0.910{\scriptsize$\pm$0.00} & 
                0.922{\scriptsize$\pm$0.00} &  
                0.905{\scriptsize$\pm$0.00}   \\
        \bottomrule
    \end{tabular}
    \setlength{\abovecaptionskip}{.5em}
    \caption{Results for different feature dissolver resolutions.}
    \label{tab:resolution_blur}
\end{table}

\section{Additional Experiments}

\subsection{Learning Anomalous Feature Patterns}
\label{sec:anomalous_feat}

This paper introduces a groundbreaking approach to fine-grained feature learning by contrasting images with their feature-dissolved counterparts. This technique enables our algorithm to identify and learn the fine-grained discriminative features for fine-grained anomaly detection. An inherited idea is to explore if our approach can enhance the detection of anomalous features by integrating a higher volume of anomalous data into the training set. As shown in~\Cref{tab:anomalous_}, there is a notable improvement in anomaly detection performance correlating with an increased presence of anomalous data.

\begin{table}[H]
    \scriptsize
    \centering
    \setlength{\tabcolsep}{5pt}
    \begin{tabular}{c|c c c c}
        \toprule
         $\lambda$ & Kvasir-Polyp & Retinal-OCT & APTOS-2019  \\
         \midrule
         $0\%$ &
            {0.860}{\scriptsize$\pm$0.04} &
            {0.944}{\scriptsize$\pm$0.01} &
            {0.934}{\scriptsize$\pm$0.00}\\
         $10\%$ &
            {0.877}{\scriptsize$\pm$0.02} &
            0.948{\scriptsize$\pm$0.01} &
            0.935{\scriptsize$\pm$0.00}
            \\
         $20\%$ & 
            \textbf{0.880}{\scriptsize$\pm$0.01} &
            \textbf{0.951}{\scriptsize$\pm$0.00} &
            \textbf{0.940}{\scriptsize$\pm$0.00}
            \\
        \bottomrule
    \end{tabular}
    \setlength{\abovecaptionskip}{.5em}
    \caption{Performance improvement with increasing proportions of anomalous data. $\lambda$ is the proportion of anomalous samples within the training data.}
    \label{tab:anomalous_}
\end{table}

\subsection{New Negative Pairs vs. Batchsize Increment}
\label{sec:abl_batch}

As the newly introduced \textit{dissolving transformation} branch, given the same batch size $B$, our proposed \textit{DIA} takes $3K\cdot B$ samples compared to the baseline \textit{CSI} that uses $2K\cdot B$ samples. In a way, \textit{DIA} increases the batchsize by a factor of $1.5$. Since contrastive learning can be batchsize dependent \cite{gutmann2012noise,he2019moco}, we demonstrate in \cref{tab:batchsize} that our performance improvement is not due to batch size. \textit{CSI} with a larger batch size exhibits similar performances as the baseline \textit{CSI} method, while the proposed \textit{DIA} method outperformed the baselines significantly.

\begin{table}[!h]
    \scriptsize
    \centering
    \begin{tabular}{
        @{\hskip7pt}l@{\hskip7pt}
        |
        @{\hskip12pt}c@{\hskip12pt}
        @{\hskip12pt}c@{\hskip12pt}
        |
        @{\hskip12pt}c@{\hskip12pt}
    }
        \toprule
        Datasets & CSI & CSI-1.5 & DIA  \\
         \midrule
         \shortstack[c]{PneumoniaMNIST} & 0.834 & 0.838 & \textbf{0.903} \\
         \shortstack[c]{BreastMNIST} & 0.546 & 0.564 & \textbf{0.750} \\
         \shortstack[c]{SARS-COV-2} & 0.785 &  0.804 & \textbf{0.851} \\
         \shortstack[c]{Kvasir-Polyp} & 0.609 & 0.679 & \textbf{0.860} \\
        \bottomrule
    \end{tabular}
    \setlength{\abovecaptionskip}{.5em}
    \caption{Comparison between DIA and the batch size increment. \textit{CSI-1.5} represents the baseline CSI models that are trained with $1.5$ times bigger batch sizes. To be specific, \textit{CSI} and \textit{DIA} are trained with a batch size of 32 while \textit{CSI-1.5} used 48.}
    \label{tab:batchsize}
\end{table}

\subsection{The Design of Similarity Matrix}
\label{sec:sim_design}

\textit{Shifting transformations} enlarge the internal distribution differences by introducing negative pairs where the views of the same image are strongly different.

\begin{figure}[h]
\begin{subfigure}{.48\linewidth}
\centering
\begin{tikzpicture}[scale=0.39]
  \foreach \y [count=\n] in {
      {100,70,70,70,0,70, 70,70,70,70,70,70},
      {70,100,70,70,70,0, 70,70,70,70,70,70},
      {70,70,100,70,70,70, 0,70,70,70,70,70},
      {70,70,70,100,70,70, 70,0,70,70,70,70},
      {0,70,70,70,100,70, 70,70,70,70,70,70},
      {70,0,70,70,70,100, 70,70,70,70,70,70},
      {70,70,0,70,70,70, 100,70,70,70,70,70},
      {70,70,70,0,70,70, 70,100,70,70,70,70},
      {70,70,70,70,70,70, 70,70,100,70,70,70},
      {70,70,70,70,70,70, 70,70,70,100,70,70},
      {70,70,70,70,70,70, 70,70,70,70,100,70},
      {70,70,70,70,70,70, 70,70,70,70,70,100},
    } {
      \ifnum \n=1
        \node[minimum size=2mm,rotate=-90] at (\n, 1.4) {\tiny $T_0(x_1)$};
      \fi
      \ifnum \n=2
        \node[minimum size=2mm,rotate=-90] at (\n, 1.4) {\tiny $T_0(x_2)$};
      \fi
      \ifnum \n=3
        \node[minimum size=2mm,rotate=-90] at (\n, 1.4) {\tiny $T_1(x_1)$};
      \fi
      \ifnum \n=4
        \node[minimum size=2mm,rotate=-90] at (\n, 1.4) {\tiny $T_1(x_2)$};
      \fi
      \ifnum \n=5
        \node[minimum size=2mm,rotate=-90] at (\n, 1.4) {\tiny $T'_0(x_1)$};
      \fi
      \ifnum \n=6
        \node[minimum size=2mm,rotate=-90] at (\n, 1.4) {\tiny $T'_0(x_2)$};
      \fi
      \ifnum \n=7
        \node[minimum size=2mm,rotate=-90] at (\n, 1.4) {\tiny $T'_1(x_1)$};
      \fi
      \ifnum \n=8
        \node[minimum size=2mm,rotate=-90] at (\n, 1.4) {\tiny $T'_1(x_2)$};
      \fi
      \ifnum \n=9
        \node[minimum size=2mm,rotate=-90] at (\n, 1.4) {\tiny ${A_0(x_1,t)}$};
      \fi
      \ifnum \n=10
        \node[minimum size=2mm,rotate=-90] at (\n, 1.4) {\tiny ${A_0(x_2,t)}$};
      \fi
      \ifnum \n=11
        \node[minimum size=2mm,rotate=-90] at (\n, 1.4) {\tiny ${A_1(x_1,t)}$};
      \fi
      \ifnum \n=12
        \node[minimum size=2mm,rotate=-90] at (\n, 1.4) {\tiny ${A_1(x_2,t)}$};
      \fi
      \foreach \x [count=\m] in \y {
        \node[fill=white!\x!blue, draw=black!20, ultra thin, minimum size=3.8mm, text=white] at (\m,-\n) {};
      }
    }
  \foreach \a [count=\i] in {
    $O_0(x_1)$,
    $O_0(x_2)$,
    $O_1(x_1)$,
    $O_1(x_2)$,
    $O'_0(x_1)$,
    $O'_0(x_2)$,
    $O'_1(x_1)$,
    $O'_1(x_2)$,
    {$A_0(x_1,t)$},
    {$A_0(x_2,t)$},
    {$A_1(x_1,t)$},
    {$A_1(x_2,t)$},
  } {
    \node[minimum size=3.8mm] at (-1.4,-\i) {\tiny \a};
  }
\end{tikzpicture}
\caption{}
\label{fig:heatmap_ablation-a}
\end{subfigure}
\begin{subfigure}{.48\linewidth}
\centering
\begin{tikzpicture}[scale=0.39]
  \foreach \y [count=\n] in {
      {100,70,70,70,0,70, 70,70,100,70,70,70},
      {70,100,70,70,70,0, 70,70,70,100,70,70},
      {70,70,100,70,70,70, 0,70,70,70,100,70},
      {70,70,70,100,70,70, 70,0,70,70,70,100},
      {0,70,70,70,100,70, 70,70,100,70,70,70},
      {70,0,70,70,70,100, 70,70,70,100,70,70},
      {70,70,0,70,70,70, 100,70,70,70,100,70},
      {70,70,70,0,70,70, 70,100,70,70,70,100},
      {100,70,70,70,100,70, 70,70,100,70,70,70},
      {70,100,70,70,70,100, 70,70,70,100,70,70},
      {70,70,100,70,70,70, 100,70,70,70,100,70},
      {70,70,70,100,70,70, 70,100,70,70,70,100},
    } {
      \ifnum \n=1
        \node[minimum size=2mm,rotate=-90] at (\n, 1.4) {\tiny $T_0(x_1)$};
      \fi
      \ifnum \n=2
        \node[minimum size=2mm,rotate=-90] at (\n, 1.4) {\tiny $T_0(x_2)$};
      \fi
      \ifnum \n=3
        \node[minimum size=2mm,rotate=-90] at (\n, 1.4) {\tiny $T_1(x_1)$};
      \fi
      \ifnum \n=4
        \node[minimum size=2mm,rotate=-90] at (\n, 1.4) {\tiny $T_1(x_2)$};
      \fi
      \ifnum \n=5
        \node[minimum size=2mm,rotate=-90] at (\n, 1.4) {\tiny $T'_0(x_1)$};
      \fi
      \ifnum \n=6
        \node[minimum size=2mm,rotate=-90] at (\n, 1.4) {\tiny $T'_0(x_2)$};
      \fi
      \ifnum \n=7
        \node[minimum size=2mm,rotate=-90] at (\n, 1.4) {\tiny $T'_1(x_1)$};
      \fi
      \ifnum \n=8
        \node[minimum size=2mm,rotate=-90] at (\n, 1.4) {\tiny $T'_1(x_2)$};
      \fi
      \ifnum \n=9
        \node[minimum size=2mm,rotate=-90] at (\n, 1.4) {\tiny ${A_0(x_1,t)}$};
      \fi
      \ifnum \n=10
        \node[minimum size=2mm,rotate=-90] at (\n, 1.4) {\tiny ${A_0(x_2,t)}$};
      \fi
      \ifnum \n=11
        \node[minimum size=2mm,rotate=-90] at (\n, 1.4) {\tiny ${A_1(x_1,t)}$};
      \fi
      \ifnum \n=12
        \node[minimum size=2mm,rotate=-90] at (\n, 1.4) {\tiny ${A_1(x_2,t)}$};
      \fi
      \foreach \x [count=\m] in \y {
        \node[fill=white!\x!blue, draw=black!20, ultra thin, minimum size=3.8mm, text=white] at (\m,-\n) {};
      }
    }
  \foreach \a [count=\i] in {
    $O_0(x_1)$,
    $O_0(x_2)$,
    $O_1(x_1)$,
    $O_1(x_2)$,
    $O'_0(x_1)$,
    $O'_0(x_2)$,
    $O'_1(x_1)$,
    $O'_1(x_2)$,
    {$A_0(x_1,t)$},
    {$A_0(x_2,t)$},
    {$A_1(x_1,t)$},
    {$A_1(x_2,t)$},
  } {
    \node[minimum size=3.8mm] at (-1.4,-\i) {\tiny \a};
  }
\end{tikzpicture}
\caption{}
\label{fig:heatmap_ablation-b}
\end{subfigure}
\setlength{\belowcaptionskip}{-1.5em}
\setlength{\abovecaptionskip}{.5em}
\caption{Visual comparison between the similarity matrices ($K=2$). The white, blue, and lavender blocks denote the excluded, positive, and negative values, respectively.}
\label{fig:heatmap_ablation}
\end{figure}
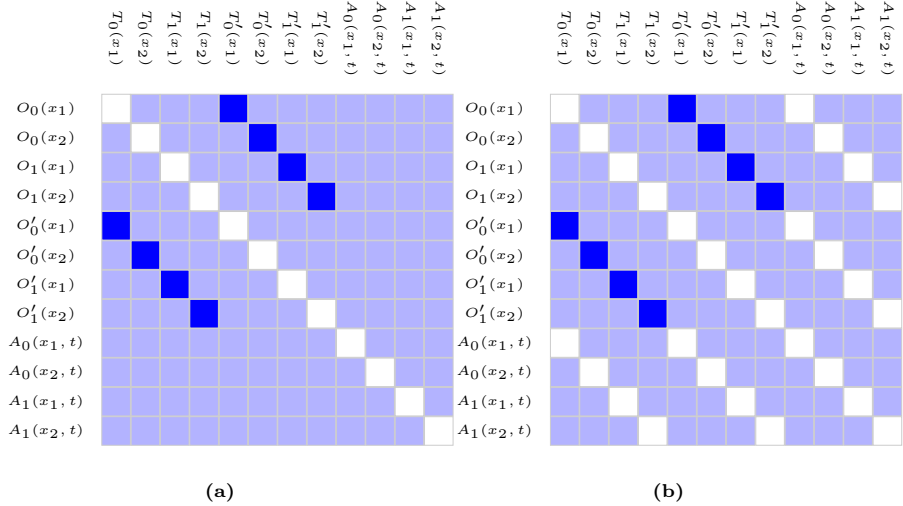

With augmentation branches $O_i$ and $O'_j$, the target similarity matrix for contrastive learning is therefore defined where the image pairs that share the same \textit{shift transformation} as positive while other combinations as negative, as presented in~\cref{fig:heatmap_ablation-a}. Due to the introduction of the dissolving transformation branch $A_k$, this ablation studies the design of the target similarity matrix of those newly introduced pairs. We further evaluate the design of \cref{fig:heatmap_ablation-b}, where the target similarity matrix is designed to exclude the image pairs with and without dissolving transformations applied whilst sharing the same \textit{shift transformation}, when $i=k$ or $j=k$. Essentially, these pairs share the same \textit{shift transformation} which should be considered as positive samples, but the $A_k$ branch removes features that make them appear negative. Thus, we investigate whether these contradictory samples should be considered during contrastive learning.
\begin{table}[h]
    \centering
    \scriptsize
    \begin{tabular}{l|cccc}
        \toprule
            Methods
            & \makecell{SARS-\\COV-2}
            & \makecell{Kvasir\\Polyp} 
            & \makecell{Retinal\\OCT}
            & \makecell{APTOS\\2019} \\
        \midrule
         {\color{darkgray} \scriptsize Baseline} \hfill CSI\;\;\;\;\;\; & 
            0.785 &
            0.609 &
            0.803 &
            {0.927} \\
        {\color{darkgray} \scriptsize Ours \hspace{2.8em}}  DIA-(a) & 
            {0.851} &
            {0.860} &
            {0.944} &
            {0.934} \\
        {\color{darkgray} \scriptsize Ours \hspace{2.8em}}  DIA-(b) &
            {0.850} &
            {0.843} &
            {0.932} &
            {0.930} \\
        \bottomrule
    \end{tabular}
    \setlength{\belowcaptionskip}{-1.5em}
    \setlength{\abovecaptionskip}{.5em}
    \caption{Semi-supervised fine-grained medical anomaly detection results.}
    \label{tab:med_app}
\end{table}

As shown in \cref{tab:med_app}, those designs achieve very similar performances on medical datasets. Then, we further evaluate our methods on standard anomaly detection datasets, that contain coarse-grained feature differences (\ie Car vs. Plane) with a minimum need to discover fine-grained features. We therefore further include the following datasets:

\textbf{{CIFAR-10}} consists of 60,000 32x32 color images in 10 equally distributed classes with 6,000 images per class, including 5,000 training images and 1,000 test images.
    
{\textbf{CIFAR-100}} similar to CIFAR-10, except with 100 classes containing 600 images each. There are 500 training images and 100 testing images per class. The 100 classes in the dataset are grouped into 20 superclasses. Each image comes with a "fine" label (the class to which it belongs) and a "coarse" label (the superclass to which it belongs), which we use in the experiments.

Note that the corresponding diffusion models for each experiment are trained on the full CIFAR10 and CIFAR100 datasets, respectively.

\begin{table}[h]
    \scriptsize
    \centering
    \begin{tabular}{
        @{\hskip3pt}c@{\hskip3pt} @{\hskip3pt}c@{\hskip3pt}|
        @{\hskip3pt}c@{\hskip3pt}
        @{\hskip3pt}c@{\hskip3pt}
        @{\hskip3pt}c@{\hskip3pt}
        @{\hskip3pt}c@{\hskip3pt}
        @{\hskip3pt}c@{\hskip3pt}
        @{\hskip3pt}c@{\hskip3pt}
        @{\hskip3pt}c@{\hskip3pt}
        @{\hskip3pt}c@{\hskip3pt}
        @{\hskip3pt}c@{\hskip3pt}
        @{\hskip3pt}c@{\hskip3pt}
        ||
        @{\hskip3pt}c@{\hskip3pt}
    }
        \toprule
        Dataset & Method & 0 & 1 & 2 & 3 & 4 & 5 & 6 & 7 & 8 & 9 & avg. \\
         \cmidrule(l){2-13}
         \multirow{3}{*}{\tiny CIFAR10}
         & {\color{darkgray} \tiny Baseline} \hfill CSI\;\;\;\;\;\; & 89.9 & 99.1 & 93.1 & 86.4 & 93.9 & 93.2 & 95.1 & 98.7 & 97.9 & 95.5 & 94.3 \\
         & {\color{darkgray} \tiny Ours \hspace{2.8em}}  DIA-(a) & 90.4 & 99.0 & 91.8 & 82.7 & 93.8 & 91.7 & 94.7 & 98.4 & 97.2 & 95.6 & 93.5 \\
         & {\color{darkgray} \tiny Ours \hspace{2.8em}}  DIA-(b) & 80.0 & 98.9 & 80.1 & 74.0 & 81.2 & 84.4 & 82.7 & 94.7 & 93.9 & 89.7 & 86.0 \\
         \midrule
         \midrule
         Dataset & Method & 0 & 1 & 2 & 3 & 4 & 5 & 6 & 7 & 8 & 9 \\
         \cmidrule(l){2-12}
         \multirow{7}{*}{\tiny CIFAR100}
         & {\color{darkgray} \tiny Baseline} \hfill CSI\;\;\;\;\;\; & 86.3 & 84.8 & 88.9 & 85.7 & 93.7 & 81.9 & 91.8 & 83.9 & 91.6 & 95.0 \\
         & {\color{darkgray} \tiny Ours \hspace{2.8em}}  DIA-(a) & 85.9 & 82.6 & 87.0 & 84.7 & 91.8 & 84.4 & 92.1 & 79.9 & 90.8 & 95.3 \\
         & {\color{darkgray} \tiny Ours \hspace{2.8em}}  DIA-(b) & 83.2 & 80.4 & 86.1 & 83.0 & 90.8 & 78.2 & 90.6 & 75.8 & 86.7 & 92.5 \\
         \cmidrule{2-13}
          & Method & 10 & 11 & 12 & 13 & 14 & 15 & 16 & 17 & 18 & 19 & avg. \\
         \cmidrule(l){2-13}
         & {\color{darkgray} \tiny Baseline} \hfill CSI\;\;\;\;\;\; & 94.0 & 90.1 & 90.3 & 81.5 & 94.4 & 85.6 & 83.0 & 97.5 & 95.9 & 95.2 & 89.6 \\
         & {\color{darkgray} \tiny Ours \hspace{2.8em}}  DIA-(a) & 93.0 & 90.1 & 89.9 & 76.7 & 93.1 & 81.7 & 79.7 & 96.0 & 96.3 & 95.2 & 88.3  \\
         & {\color{darkgray} \tiny Ours \hspace{2.8em}}  DIA-(b) & 91.2 & 86.3 & 87.7 & 73.3 & 91.8 & 80.7 & 79.7 & 97.2 & 95.3 & 93.3 &86.2 \\
        \bottomrule
    \end{tabular}
    \setlength{\belowcaptionskip}{-1.5em}
    \setlength{\abovecaptionskip}{.5em}
    \caption{Results on standard benchmark datasets.
    Results are AUROC scores that are scaled by 100.
    }
    \label{tab:std_app}
\end{table}

As shown in \cref{tab:med_app} and \cref{tab:std_app}, the exclusion of the $i=k$ and $j=k$ pairs barely affect the performance for the fine-grained anomaly detection tasks, but significantly lowers the performance for the coarse-grained anomaly detection tasks.

\subsection{Memory footprint}

The computational efficiency is provided in \Cref{tab:complexity}. We provide the memory footprint as below:

\begin{table}[h]
    \centering
    \begin{tabular}{ccccc}
        \toprule
        Batch size 
            & 8 
            & 16 
            & 32
            & 64  \\
        \midrule
         GPU mem (GB) & 2.38 & 4.51 & 8.78 & 17.33 \\
        \bottomrule
    \end{tabular}
    \setlength{\belowcaptionskip}{-1.5em}
    \setlength{\abovecaptionskip}{.5em}
    \caption{Memory footprint on different image resolutions.}
    \label{tab:memory}
\end{table}

\section{Non-Data-Specific Dissolving}
\label{sec:dslv-sd}
As per the discussion in~\cref{sec:role,sec:discussion}, we demonstrated the importance of the training for data-specific diffusion models. To further provide an intuition of what happens when using non-data-specific diffusion models, we present visual examples for the dissolving transformations with ``incorrect" models. For each dataset, we show the expected dissolved images using the data-specific diffusion models (as used in our framework), dissolving with a diffusion model trained on PneumoniaMNIST dataset, dissolving with a diffusion model trained on CIFAR10 dataset, and dissolving with Stable Diffusion\footnote{Stable diffusion performs reverse diffusion steps on the latent feature space. We, therefore, use the VAE model to encode the image to latent space for the dissolving transformation. Then we decode the latent features back to images.}~\cite{rombach2021highresolution}.

\begin{figure*}
    \includegraphics[width=\linewidth,trim={0 0 .88cm 0},clip]{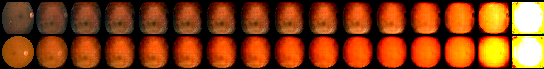}
    \includegraphics[width=\linewidth]{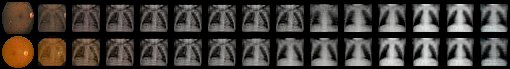}
    \includegraphics[width=\linewidth,trim={0 0 4.3cm 0},clip]{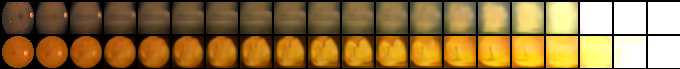}
    \includegraphics[width=\linewidth]{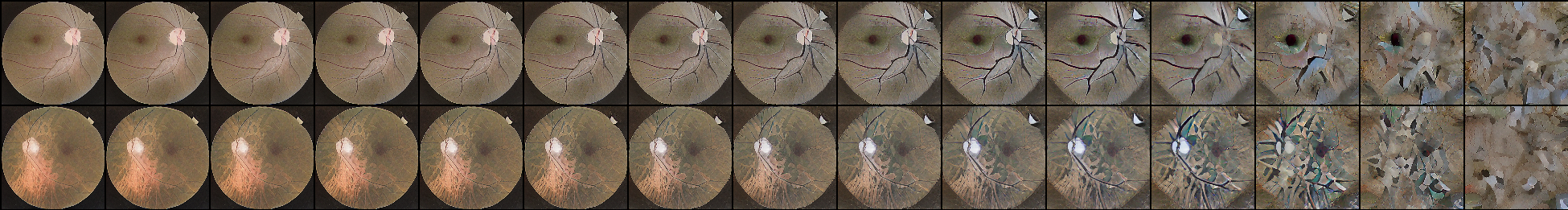}

    \setlength{\belowcaptionskip}{-1.5em}
    \setlength{\abovecaptionskip}{.5em}
    \caption{Visualization of APTOS dataset. From left to right are the dissolved images with increased $t$ from 1 to 975. From top to bottom, the first three rows represent models trained on the APTOS, PneumoniaMNIST, and CIFAR10 datasets, respectively. The final row showcases the output of the stable diffusion model.}
    \label{fig:dslv_aptos}
\end{figure*}

\begin{figure*}
    \includegraphics[width=\linewidth,trim={0 0 .88cm 0},clip]{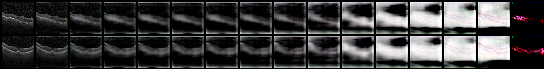}
    \includegraphics[width=\linewidth]{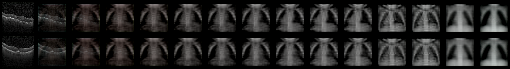}
    \includegraphics[width=\linewidth,trim={0 0 4.3cm 0},clip]{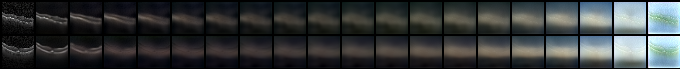}
    \includegraphics[width=\linewidth]{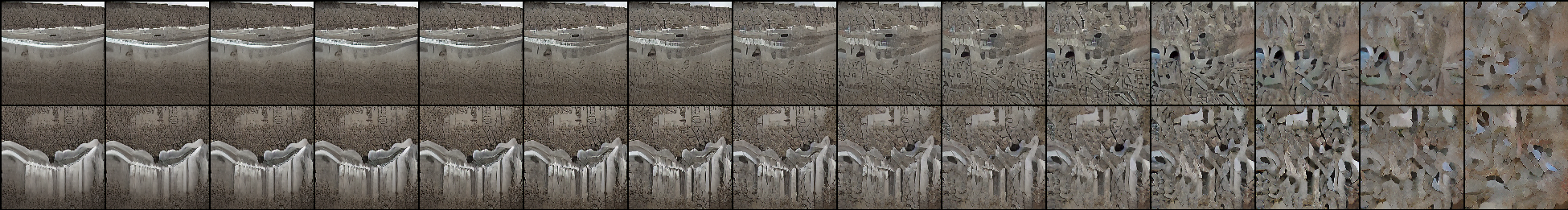}
    \setlength{\belowcaptionskip}{-1.5em}
    \setlength{\abovecaptionskip}{.5em}
    \caption{Visualization of OCT2017 dataset. From left to right are the dissolved images with increased $t$ from 1 to 975. From top to bottom, the first three rows represent models trained on the OCT2017, PneumoniaMNIST, and CIFAR10 datasets, respectively. The final row showcases the output of the stable diffusion model.}
    \label{fig:dslv_oct}
\end{figure*}

\begin{figure*}
    \includegraphics[width=\linewidth,trim={0 0 .88cm 0},clip]{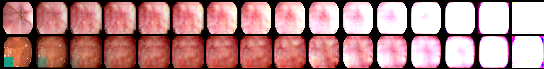}
    \includegraphics[width=\linewidth]{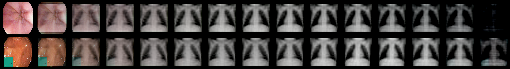}
    \includegraphics[width=\linewidth,trim={0 0 4.3cm 0},clip]{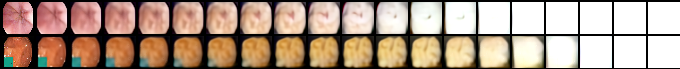}
    \includegraphics[width=\linewidth]{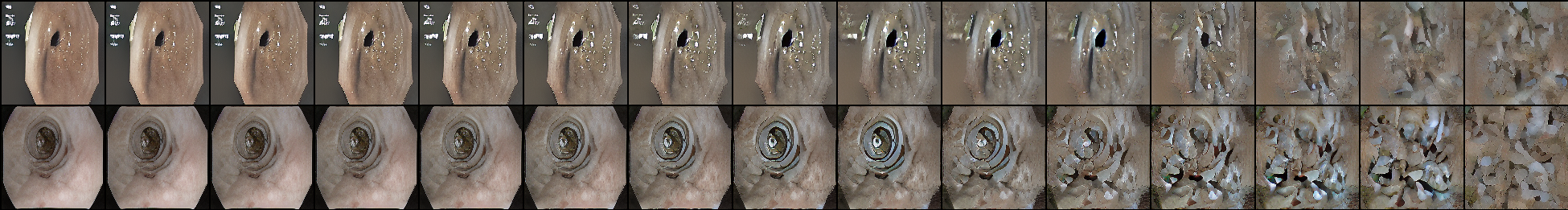}
    \caption{Visualization of Kvasir dataset. From left to right are the dissolved images with increased $t$ from 1 to 975. From top to bottom, the first three rows represent models trained on the Kvasir, PneumoniaMNIST, and CIFAR10 datasets, respectively. The final row showcases the output of the stable diffusion model.}
    \label{fig:dslv_kvasir}
\end{figure*}

\begin{figure*}
    \includegraphics[width=\linewidth,trim={0 0 .88cm 0},clip]{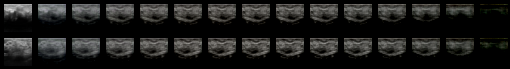}
    \includegraphics[width=\linewidth]{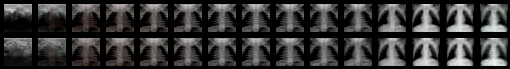}
    \includegraphics[width=\linewidth,trim={0 0 4.3cm 0},clip]{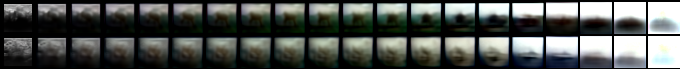}
    \includegraphics[width=\linewidth]{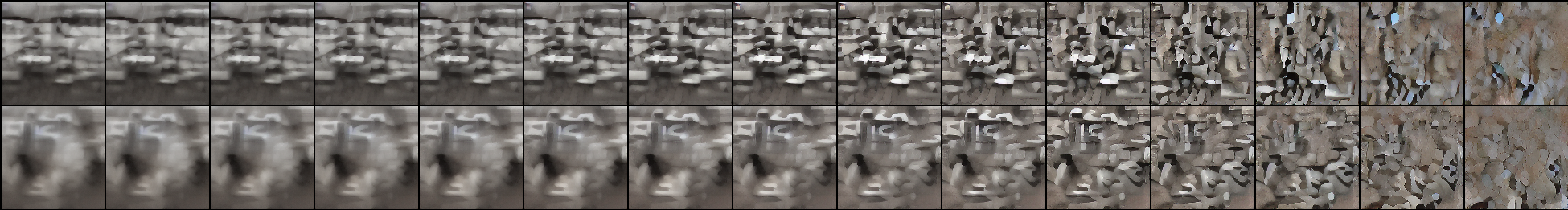}
    \setlength{\belowcaptionskip}{-1.5em}
    \setlength{\abovecaptionskip}{.5em}
    \caption{Visualization of BreastMNIST dataset. From left to right are the dissolved images with increased $t$ from 1 to 975. From top to bottom, the first three rows represent models trained on the BreastMNIST, PneumoniaMNIST, and CIFAR10 datasets, respectively. The final row showcases the output of the stable diffusion model.}
    \label{fig:dslv_breast}
\end{figure*}

\begin{figure*}[t!]
    \includegraphics[width=\linewidth,trim={0 0 .82cm 0},clip]{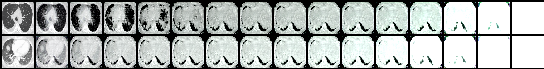}
    \includegraphics[width=\linewidth]{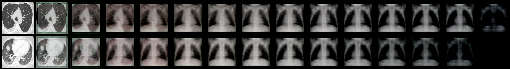}
    \includegraphics[width=\linewidth,trim={0 0 4.28cm 0},clip]{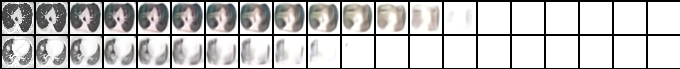}
    \includegraphics[width=\linewidth]{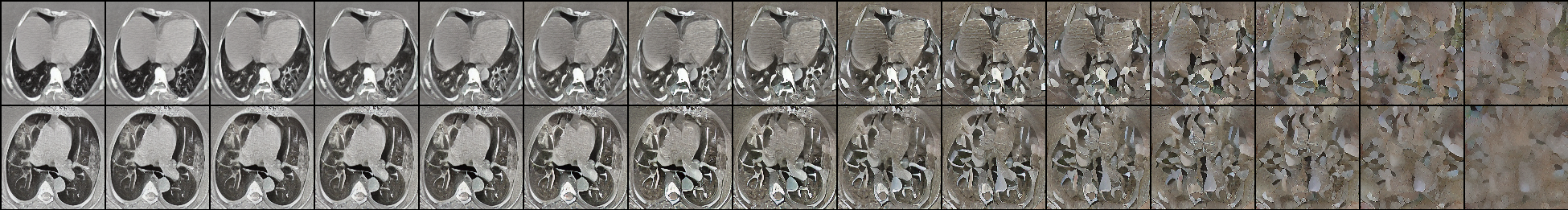}
    \caption{Visualization of SARS-COVID-2 dataset. From left to right are the dissolved images with increased $t$ from 1 to 975. From top to bottom, the first three rows represent models trained on the SARS-CoV-2, PneumoniaMNIST, and CIFAR10 datasets, respectively. The final row showcases the output of the stable diffusion model.}
    \vspace*{5in}
    \label{fig:dslv_sars}
\end{figure*}

As illustrated in~\cref{fig:dslv_aptos,fig:dslv_oct,fig:dslv_kvasir,fig:dslv_breast,fig:dslv_sars}, the dissolving operation dissolves images towards the learned prior of the training dataset. Such behavior is especially significant by using the PneumoniaMNIST trained diffusion model. We can observe that all images soon look like lung x-rays, regardless of how the input looks like. For the Stable Diffusion model, the dissolving transformation removes the texture and then corrupts the image.



\end{document}